\documentclass[dvipsnames]{article}

\usepackage[final]{delta_tuning}

\usepackage[utf8]{inputenc} 
\usepackage[T1]{fontenc}    
\usepackage{url}            
\usepackage{booktabs}       
\usepackage{amsfonts}       
\usepackage{nicefrac}       
\usepackage{microtype}      
\usepackage{epigraph}
\usepackage{enumitem}
\usepackage{graphicx}
\usepackage{multirow}
\usepackage{multicol}
\usepackage{amsmath}
\usepackage{mathrsfs}
\usepackage{textcomp}
\usepackage{subcaption}

\usepackage{array}
\usepackage{color, soul}
\usepackage[most,breakable]{tcolorbox}
\usepackage{xcolor}        

\usepackage[labelfont=bf]{caption}

\usepackage[colorlinks, linkcolor=RoyalBlue, anchorcolor=BrickRed, citecolor=RoyalBlue, urlcolor=RoyalBlue]{hyperref}

\title{Configurable Foundation Models: \\ Building LLMs from a Modular Perspective}

\author{
Chaojun Xiao$^1$, Zhengyan Zhang$^1$, Chenyang Song$^1$, Dazhi Jiang$^1$, Feng Yao$^2$, Xu Han$^1$\thanks{ Corresponding authors.}\hspace{0.5em} \vspace{0.03in}\\ \textbf{Xiaozhi Wang$^1$, Shuo Wang$^1$, Yufei Huang$^1$, Guanyu Lin$^3$, Yingfa Chen$^1$, Weilin Zhao$^1$} \vspace{0.03in}\\ \textbf{Yuge Tu$^4$, Zexuan Zhong$^6$, Ao Zhang$^7$, Chenglei Si$^8$,
Khai Hao Moo$^4$, Chenyang Zhao$^9$
} \vspace{0.03in}\\ \textbf{Huimin Chen$^1$, Yankai Lin$^5$, Zhiyuan Liu$^{1,4*}$, Jingbo Shang$^2$, Maosong Sun$^{1*}$ } \vspace{0.1in}\\ 
$^1$Tsinghua University, $^2$University of California San Diego, $^3$Carnegie Mellon University \vspace{0.03in}\\ $^4$ModelBest Inc., $^5$Renmin University of China, $^6$Princeton University \vspace{0.03in}\\ $^7$National University of Singapore, $^8$Stanford University, $^9$University of California, Los Angeles\vspace{0.03in} \\
\texttt{xiaocj20@mails.tsinghua.edu.cn}\\
\texttt{\{hanxu2022,liuzy,sms\}@tsinghua.edu.cn}
}

\begin{document}

\setlength\epigraphrule{0pt}
\setlength\epigraphwidth{.7\textwidth}

\maketitle

\begin{abstract}
Advancements in large language models (LLMs) have recently unveiled challenges tied to computational efficiency and continual scalability due to their requirements of huge parameters, making the applications and evolution of these models on devices with limited computation resources and scenarios requiring various abilities increasingly cumbersome. Inspired by modularity within the human brain, there is a growing tendency to decompose LLMs into numerous functional modules, allowing for inference with part of modules and dynamic assembly of modules to tackle complex tasks, such as mixture-of-experts. To highlight the inherent efficiency and composability of the modular approach, we coin the term \textit{brick} to represent each functional module, designating the modularized structure as \textit{configurable foundation models}. In this paper, we offer a comprehensive overview and investigation of the construction, utilization, and limitation of configurable foundation models. We first formalize modules into \textit{emergent bricks} - functional neuron partitions that emerge during the pre-training phase, and \textit{customized bricks} - bricks constructed via additional post-training to improve the capabilities and knowledge of LLMs. 
Based on diverse functional bricks, we further present four brick-oriented operations: retrieval and routing, merging, updating, and growing. These operations allow for dynamic configuration of LLMs based on the instruction to handle complex tasks.
To verify our perspective, we conduct an empirical analysis on widely-used LLMs, Llama-3-8B-Instruct and Mistral-7B-Instruct-v0.3. We find that the FFN layers follow modular patterns with functional specialization of neurons and functional neuron partitions.
Finally, as the domain of configurable LLMs remains nascent and evolving, we highlight several open issues and directions for future research, including the correlation between emergent and customized bricks, general brick development protocols, evaluation of configurable LLMs, efficient brick computing frameworks, and systems consisting of multiple model-level bricks.
Overall, this paper aims to offer a fresh modular perspective on existing LLM research and inspire the future creation of more efficient and scalable foundational models.

\end{abstract}

\hspace{80pt}\parbox[b]{0.45\textwidth}
{
\epigraph{\textit{``Rome was not built in a day, but they were laying bricks every hour.''}}{--- John Heywood}
}

\newpage
{
  \hypersetup{linkcolor=RoyalBlue, linktoc=page}
  \tableofcontents
}

\newpage

\section{Introduction}

Large pre-trained models, especially large pre-trained language models (LLMs), have achieved remarkable success in a variety of tasks~\citep{DBLP:conf/naacl/DevlinCLT19,DBLP:conf/nips/BrownMRSKDNSSAA20,DBLP:journals/aiopen/HanZDGLHQYZZHHJ21,DBLP:journals/corr/abs-2303-08774,DBLP:journals/corr/abs-2307-09288,DBLP:journals/corr/abs-2303-18223}. LLMs have become the foundation models of artificial intelligence applications by providing amounts of world knowledge~\citep{DBLP:conf/emnlp/PetroniRRLBWM19,shin2020autoprompt} and powerful reasoning capabilities~\citep{DBLP:journals/corr/abs-2108-07258}. 
Current advanced LLMs, such as GPT-4~\citep{DBLP:journals/corr/abs-2303-08774}, are deployed on large-scale central servers with high-bandwidth memory and GPUs to address various user instructions. With the development of LLMs, the future applications of LLMs will inevitably face the following trends, which in turn present challenges for LLMs:

(1)~\textbf{Deployment on end devices}. With the capabilities of LLMs continuing to improve, the trend of deploying these models on devices with limited computing power, such as smartphones and personal computers, is attracting increasing attention, allowing LLMs to serve as personal assistants for millions of users~\citep{apple-intelligence,xue2024powerinfer,hu2024minicpm}. The use of monolithic LLMs that require substantial computational resources is gradually becoming infeasible, and improving the computational efficiency of LLMs is a significant challenge.
(2)~\textbf{Widespread application across multiple domains}. LLMs are widely applied in various fields and applications to enhance people's work efficiency~\citep{kaddour2023challenges}. However, the knowledge and capabilities required by different domains, users, and even different instructions vary greatly. Storing all world knowledge in monolithic LLMs and serving all scenarios with full parameters often leads to redundant computations, and conflicts between different domain knowledge may even result in sub-optimal performance.
(3)~\textbf{Rapid evolution in new scenarios}. As application scenarios increase and time progresses, we usually need LLMs to efficiently adapt to new tasks and learn from environment feedbacks~\citep{li2024generative,tao2024survey}. Meanwhile, the world knowledge stored in LLMs is constantly updating and expanding. This demands that LLMs are able to evolve efficiently and continuously, learning new knowledge and skills while avoiding forgetting existing knowledge.

\begin{figure}[b]
    \centering\includegraphics[width=\textwidth]{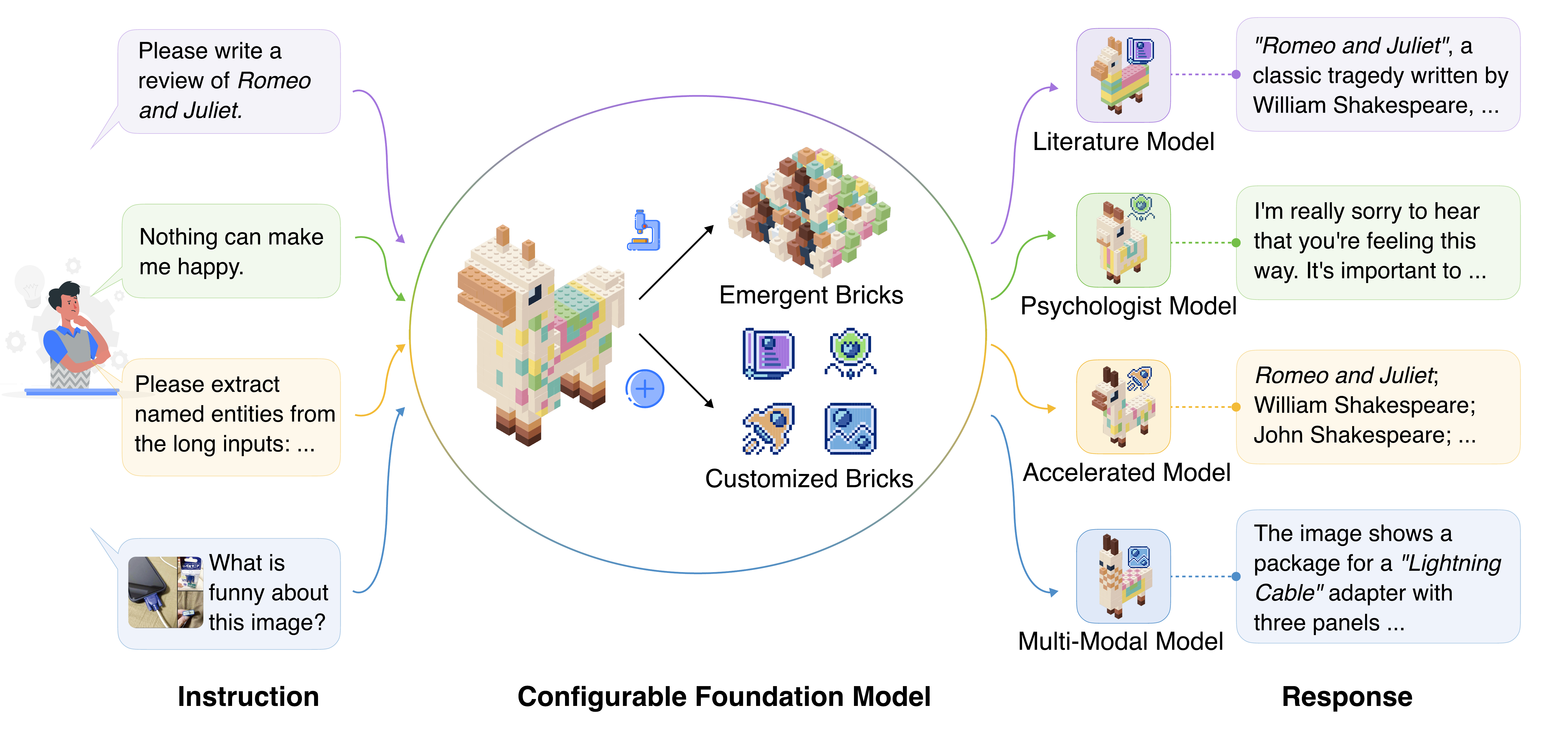}
    \caption{The illustration of a configurable foundation model consisting of emergent and customized bricks. For a given instruction, we select and combine tiny bricks to build an efficient instruction-specific model with minimal performance loss.}
    \label{fig:illustration}
\end{figure}

To address these issues, studying and analyzing LLMs from a modular perspective has gradually become an important focus for current researchers~\citep{DBLP:journals/corr/abs-2302-11529,DBLP:journals/corr/abs-2209-01667,DBLP:journals/natmi/DingQYWYSHCCCYZWLZCLTLS23,DBLP:conf/acl/ZhangZLXW00XS023}. These works decompose LLMs into functional modules. In this way, for each computation step, we can only involve part of modules to save computation costs and achieve efficient continual training by constructing or updating modules.
Modularity has long been an endogenous mechanism or a central principle in diverse fields, ranging from biomedical sciences to engineering fields~\citep{simon1962architecture}. A module is conceived as an independent component with specialized functionality that can coordinate and interface with other modules, thereby giving rise to complex systemic behavior. Owing to modularity, many complex systems become more understandable and scalable. For instance, in cognitive neuroscience, the modularity of mind hypothesis posits that the human brain comprises functionally specialized modules, such as the visual cortex for processing visual inputs~\citep{grill2004human} and Broca's area for speech production~\citep{fodor1983modularity}. These modules operate on distinct types of information while collaborating to generate integrated cognition and behavior. In software engineering, decomposing programs into logical modules can significantly enhance development efficiency, reduce project complexity, and encourage code reuse~\citep{DBLP:journals/aim/StefikB86}. In industrial manufacturing, items like electrical appliances and automobiles are also produced by assembling modular components~\citep{bejan2006constructal}.

The core characteristics of modularity are independence, specificity, and composability. Modules are decoupled from each other and can thus be specified to focus on certain functionality. By constructing and composing modules, complex systems can be built and maintained more easily. Owing to modularity, it becomes possible to simplify the systems and enhance their efficiency. It also allows for a streamlined process in developing and updating systems, ensuring they remain adaptable and scalable over time. 
Inspired by existing observations in other disciplines, modularity is becoming a promising conceptual perspective for designing and analyzing the next generation of foundation models~\citep{zador2022toward}. Some preliminary efforts indicate that LLMs have the potential to be decomposed into various specialized functional modules.
For example, we can find that LLMs adopt single neurons as memory modules to store a specific structured knowledge triplet like (\textit{China}, \texttt{Capital}, \textit{Beijing}) (i.e., Beijing is the capital of China)~\citep{DBLP:conf/acl/DaiDHSCW22,DBLP:journals/tacl/SajjadDD22,DBLP:conf/nips/MengBAB22}. Recent efforts in parameter-efficient tuning have demonstrated that constructing modules with several dozen neurons can equip LLMs with specific task abilities~\citep{DBLP:journals/natmi/DingQYWYSHCCCYZWLZCLTLS23,DBLP:conf/iclr/HuSWALWWC22,DBLP:conf/icml/HoulsbyGJMLGAG19,DBLP:conf/acl/LiL20}. To augment LLMs with multi-modal processing capabilities, some researchers attempt to adopt visual models as the modules of LLMs to analyze visual semantics~\citep{DBLP:conf/nips/AlayracDLMBHLMM22,DBLP:journals/corr/abs-2305-08848}. \textbf{Therefore, building LLMs with modules from the neuron level to the model level can significantly enhance the abilities of LLMs without costly re-training from scratch}.
Generally, a module comprises a collection of function-specific neurons, the size of which can vary according to functional requirements. In some cases, we can even regard entire pre-trained models as a functional module, beyond the definition of a module in the general sense. To avoid ambiguity and more intuitively show the benefits of modularity for LLMs, \textbf{we term the functional components to build LLMs as bricks} instead of modules. 

\textbf{Training and deploying LLMs from the perspective of brick combination enable configurable model usage}. As shown in Figure~\ref{fig:illustration}, for a given instruction, unlike computing the monolithic LLMs, we can select part of bricks with specific functionalities for computation according to the instruction functionality requirements~\citep{DBLP:journals/corr/abs-2306-04640,DBLP:conf/acl/ZhangZLXW00XS023}, which significantly benefits the computational efficiency. Besides, the adaptation and augmentation of LLMs can be further formalized as the problem of constructing new bricks or updating existing bricks~\citep{DBLP:conf/naacl/GururanganLHSZ22,DBLP:conf/emnlp/CaoAT21,DBLP:conf/acl/XiaoZHCLLLLCS23,DBLP:conf/acl/ZhangZLWYXHLLSZ23}, which are more cost-efficient and scalable than continual training of full models. \textbf{Owing to the high scalability brought by brick combination, we term the LLMs built on bricks as configurable foundation models.}
These beneficial endeavors can well propel the utilization of LLMs in daily applications, thereby offering fresh insights and approaches for the development of next-generation architectures for foundation models.
Therefore, to facilitate the progress of configurable foundation models, this paper places its emphasis on a comprehensive analysis of existing efforts, future directions, and potential challenges.

To take advantage of configurable foundation models, this paper focuses on addressing two problems: 

\textbf{\textit{Problem 1: how can we formalize and construct bricks for configurable foundation models?} ($\S\ \ref{framework}$)} 

\textbf{From the micro parameter level rather than the macro model level, both pre-training and post-training essentially involve constructing bricks for LLMs}. In recent years, the prevailing paradigm for building LLMs involves two steps: pre-training and post-training. Here, post-training includes fine-tuning and preference learning. During the pre-training phase, LLMs acquire versatile knowledge and learn general language reasoning through self-supervised learning on massive unsupervised data. During the subsequent post-training phase, LLMs are adapted to obtain additional capabilities, including downstream task ability, and domain knowledge~\citep{DBLP:journals/corr/abs-2108-07258}. Despite the pre-training and post-training processes being conducted on monolithic LLMs, recent works indicate that the impact of these processes on the internal parameters of LLMs tends to be modular~\citep{DBLP:conf/acl/ZhangZLXW00XS023,DBLP:conf/acl/AghajanyanGZ20}. 

\textbf{Pre-training is to construct \textit{emergent} bricks, the functional specialization of which emerges during pre-training. ($\S\ \ref{emergent}$)}
It includes both dense and sparse pre-training processes.
(1)~Prior analyses on the typical \textit{dense} pre-training of LLMs reveal that model parameters undergo differentiation throughout the process~\citep{DBLP:conf/acl/DaiDHSCW22,DBLP:conf/emnlp/ForoutanBLBA22,dobs2022brain,DBLP:conf/acl/ZhangZLXW00XS023}.
This dynamic and implicit process leads to the emergence of functional partitions, giving rise to an implicit brick structure. Notably, \citet{DBLP:conf/acl/ZhangZLXW00XS023} identify distinct groups of parameters within LLMs dedicated to semantic, knowledge, and task functions. \citet{DBLP:conf/emnlp/ForoutanBLBA22} uncover the presence of language-neutral sub-networks for multilingual foundation models.
Moreover, using the pruning technique, previous works have explored to discover task-specific subnetworks from LLMs~\citep{DBLP:conf/iclr/FrankleC19,wang-etal-2020-structured,xia-etal-2022-structured}.
(2)~Some recent efforts have attempted to build LLMs with the \textit{sparse} mixture-of-expert (MoE) structure~\citep{DBLP:conf/iclr/ShazeerMMDLHD17,DBLP:conf/iclr/LepikhinLXCFHKS21,DBLP:journals/jmlr/FedusZS22,DBLP:journals/corr/abs-2209-01667}. 
In the MoE structure, an LLM is composed of multiple experts, each of which has the same architecture as the feed-forward network or the attention network in the original model.
Only a subset of experts is activated at a time by a gating network to process input data.
While the primary intent of MoE methods is to enhance model capacity without escalating computational cost, these methods invisibly formalize LLMs into a pre-defined structure akin to combining bricks, where each expert functions as a distinct and specialized brick. Further studies show that the MoE structure can yield comparable results to conventional dense structures~\citep{mistralmoe} and even prove more advantages for understanding task-specific instructions due to the functional specialization~\citep{shen2023mixtureofexperts}. 
Based on such evidence, whether we intentionally or unintentionally design the structure of LLMs, bricks are spontaneously formed to target specific functions during the pre-training process.

\textbf{Post-training essentially is to construct additional \textit{customized} bricks for the whole model, the functional specialization of which is manually defined to meet specific human-defined requirements. ($\S\ \ref{plugin}$)}
To enhance models with additional abilities, such as domain knowledge and task-specific capabilities, traditional methods involve full-parameter fine-tuning. Recent research shows that parameter changes are intrinsically low-rank~\citep{DBLP:conf/iclr/LiFLY18,DBLP:conf/acl/AghajanyanGZ20,DBLP:journals/corr/abs-2110-07867}, which implies that only a small proportion of the parameters necessitate tuning for further capabilities adaptation. Inspired by these findings, parameter-efficient fine-tuning (PEFT) freezes LLMs and introduces extra parameters to achieve efficient task adaptation~\citep{DBLP:conf/iclr/HeZMBN22,DBLP:journals/natmi/DingQYWYSHCCCYZWLZCLTLS23,DBLP:journals/csur/LiuYFJHN23}. Beyond PEFT, many studies find that the additional parameters can not only endow LLMs with task-specific capabilities, but also supplement them with much extra knowledge and functionalities
, such as knowledge bricks for world knowledge injection~\citep{DBLP:conf/acl/ZhangZLWYXHLLSZ23,DBLP:conf/acl/XiaoZHCLLLLCS23}, modality bricks for multi-modal composition~\citep{DBLP:conf/nips/AlayracDLMBHLMM22,DBLP:conf/icml/0008LSH23}, memory bricks for long-text processing~\citep{xiao2024infllm}, and compression bricks for inference acceleration~\citep{DBLP:conf/emnlp/XiaoLZZHLZXL0Z23}. 
Therefore, the essence of fine-tuning LLMs is to customize bricks, which can fully supplement and stimulate knowledge and capabilities for LLMs to meet specific requirements.
Furthermore, each LLM itself can also become a customized brick in a multi-model system. For example, in a multi-agent system, each model is responsible for a specific sub-task~\citep{DBLP:journals/corr/abs-2308-11432,DBLP:journals/corr/abs-2308-10848}; in a combination of multi-modal models, each model is tasked with processing data from a specific modality~\citep{DBLP:conf/nips/AlayracDLMBHLMM22}.

After formalizing pre-training and fine-tuning into constructing bricks, we delve into a discussion on selecting brick granularity for building configurable foundation models. This entails a thoughtful evaluation of both brick capabilities and brick management, highlighting the relationship between the granularity and the capability complexity and the inclusion among different bricks ($\S\ \ref{granularity}$). 
Subsequently, we further summarize five major advantages of configurable foundation models, including efficiency, reusability, traceability, sustainability, and distributed computation ($\S\ \ref{benefit}$).

\begin{table}[t]
    \setlength{\extrarowheight}{2pt}
    \centering
    \small
    \caption{The definition of concepts coined in this paper.}
    \begin{tabular}{l|p{9.5cm}}
    \toprule
    Concept          & Definition \\ \midrule
    \textbf{Configurable Foundation Model} 
            & Refers to a foundation model composed of multiple functional components, derived from decomposing LLMs. Upon receiving specific instructions, a configurable foundation model can dynamically select and combine these components as required for compositional abilities, thereby offering configurable functionality. \\
    \textbf{Brick} 
            & Refers to a functional unit composed of a group of function-specific neurons. Its size can vary from a single neuron to even an entire model. Compared to the traditional definition of ``module'', the brick demonstrates a richer granularity and functionality. \\
    \textbf{Emergent Brick} 
            & Refers to a functional brick formed within an LLM during the pre-training process, where randomly initialized parameters gradually evolve and differentiate into specific functions. Parameters with similar functions constitute these emergent bricks. \\
    \textbf{Customized Brick (Plugin)} 
            & Refers to a functional brick specifically designed and trained to meet the needs of downstream applications. These bricks or plugins can be tailored for specific tasks, knowledge graphs, or various modalities. \\
    \bottomrule
    \end{tabular}
    \label{tab:concept}
    \vspace{-1em}
\end{table}

\begin{table*}[]
    \centering
    \small
    \caption{Existing representative efforts for configurable foundation models.}
    \begin{tabular}{l|p{1.6cm}|p{10.5cm}}
    \toprule
    \multicolumn{2}{l|}{Topic}  & Representative Reference \\ \midrule
    \multirow{27}{*}{Architecture}
        & \multirow{12}{1.6cm}{Emergent Bricks}
            & \textbf{Activation Sparsity}: Activation Sparsity of Neurons~\citep{DBLP:conf/acl/ZhangL00S022,DBLP:conf/iclr/LiYBLRRYCYGK23,zhang2024relu}, Improving the Sparsity~\citep{song2024prosparse,song2024turbo}, Inference Infrastructure with Sparsely Activated Neurons~\citep{song2023powerinfer,xue2024powerinfer,alizadeh2023llm} \\
           && \textbf{Function Localization}: Knowledge~\citep{DBLP:conf/acl/DaiDHSCW22,DBLP:conf/nips/MengBAB22}, Skill~\citep{DBLP:conf/emnlp/WangWZ0LL22,DBLP:journals/corr/abs-2309-12263,DBLP:conf/icml/PanigrahiSZA23}, Linguistic~\citep{DBLP:journals/corr/abs-2305-01610,DBLP:journals/corr/abs-2309-04827,tang2024language,zhao2023unveiling} \\
           && \textbf{Human-Defined Emergent Bricks}: Human-Defined Layers~\citep{rogers-etal-2020-primer,DBLP:conf/emnlp/GevaSBL21}, Mixture-of-Expert~\citep{DBLP:conf/iclr/ShazeerMMDLHD17,DBLP:journals/jmlr/FedusZS22} \\
           && \textbf{Self-Organized Emergent Bricks}: Neuron Grouping in FFNs~\citep{DBLP:conf/acl/ZhangL00S022,DBLP:journals/corr/abs-2310-04361,DBLP:conf/acl/ZhangZLXW00XS023,dong2024prompt} \\
           \cmidrule{2-3}
        & \multirow{14}{1.6cm}{Customized Bricks}
            & \textbf{Intrinsic Dimensionality}: Intrinsic Dimension of LLM Fine-tuning~\citep{DBLP:conf/acl/AghajanyanGZ20,DBLP:journals/corr/abs-2110-07867,DBLP:conf/acl/ZhangLS23} \\
           && \textbf{Task Bricks}: Adapter~\citep{DBLP:conf/icml/HoulsbyGJMLGAG19}, Prompt~\citep{DBLP:journals/csur/LiuYFJHN23,DBLP:conf/emnlp/LesterAC21}, Prefix Tuning~\citep{DBLP:conf/acl/LiL20}, LoRA~\citep{DBLP:conf/iclr/HuSWALWWC22}, BitFit~\citep{DBLP:conf/acl/ZakenGR22}, Steering Vector~\citep{DBLP:journals/corr/abs-2310-01405,DBLP:journals/corr/abs-2311-06668} \\
           && \textbf{Knowledge Bricks}: Knowledge Graph Brick~\citep{DBLP:conf/acl/ZhangZLWYXHLLSZ23,DBLP:conf/emnlp/PornerWS20,zhang2024kb}, Document Brick~\citep{DBLP:conf/acl/XiaoZHCLLLLCS23,gim2024prompt} \\
           && \textbf{Modality Bricks}: Modality Bricks with Textual Interface~\citep{DBLP:journals/corr/abs-2305-06355,DBLP:journals/corr/abs-2303-04671,DBLP:journals/corr/abs-2303-17580,DBLP:journals/corr/abs-2303-11381}, Modality Bricks with Continuous Interface~\citep{DBLP:conf/nips/AlayracDLMBHLMM22,DBLP:conf/icml/0008LSH23,DBLP:journals/corr/abs-2304-08485,DBLP:journals/corr/abs-2304-10592,DBLP:journals/corr/abs-2305-15023,DBLP:journals/corr/abs-2309-05519} \\
           && \textbf{Other Customized Bricks}: Bricks for Tool Using~\citep{DBLP:journals/corr/abs-2301-12652,DBLP:conf/acl/YuXY023,DBLP:journals/corr/abs-2305-11554}, Bricks for Debiasing~\citep{DBLP:conf/iclr/DathathriMLHFMY20}, Bricks for Acceleration~\citep{DBLP:conf/emnlp/XiaoLZZHLZXL0Z23}, Bricks for Style Transfer~\citep{DBLP:conf/emnlp/PascualEMCW21,DBLP:conf/iclr/DathathriMLHFMY20} \\
    \midrule
    \multirow{27}{*}{Operation}
        & \multirow{4}{1.6cm}{Routing and Retrieval}
            & \textbf{Routing for Emergent Brick}: Trainable Routing~\citep{DBLP:journals/jmlr/FedusZS22,DBLP:conf/icml/LewisBDGZ21,DBLP:conf/nips/ZhouLLDHZDCLL22,DBLP:journals/corr/abs-2308-00951,qiu2024layerwise}, Fixed Router~\citep{DBLP:conf/nips/RollerSSW21,DBLP:conf/iclr/Zuo00KHZGZ22,DBLP:conf/naacl/GururanganLHSZ22} \\
           && \textbf{Retrieval for Customized Brick}: Knowledge Brick Retrieval~\citep{DBLP:conf/emnlp/FriedmanDC21,DBLP:conf/eacl/PfeifferKRCG21, DBLP:journals/corr/abs-2307-13269}, Task Brick Retrieval~\citep{DBLP:journals/corr/abs-2402-09997} \\
        \cmidrule{2-3}
        & \multirow{8}{*}{Combination}
            & \textbf{Parameter Weighted Averaging}: Ensemble of Bricks with the Same Ability~\citep{DBLP:conf/nips/GaripovIPVW18,DBLP:conf/icml/WortsmanIGRLMNF22,DBLP:conf/emnlp/QinQYCLHLSZ22,DBLP:conf/eacl/ChronopoulouPFD23,DBLP:conf/iclr/RuanSMAIFD23,DBLP:conf/nips/ArpitWZX22,DBLP:conf/nips/RameKRRGC22,DBLP:journals/corr/abs-2208-03306}, Composition of Bricks with Different Abilities~\citep{DBLP:conf/nips/MatenaR22,DBLP:conf/iclr/Jin0P023,DBLP:conf/iclr/IlharcoRWSHF23,DBLP:journals/corr/abs-2306-14870,DBLP:journals/corr/abs-2303-17574} \\
           && \textbf{Brick Stitching}: Heuristic Stitching~\citep{DBLP:conf/nips/AlayracDLMBHLMM22,DBLP:conf/icml/0008LSH23,DBLP:journals/corr/abs-2308-11432,DBLP:conf/cvpr/Pan0Z23,akiba2024evolutionary}, Planner-based Stitching~\citep{DBLP:conf/cvpr/AndreasRDK16,DBLP:conf/iccv/HuARDS17,DBLP:journals/ijon/Fashandi23,DBLP:conf/iclr/YaoZYDSN023,DBLP:journals/corr/abs-2306-08640}   \\
        \cmidrule{2-3}
        & \multirow{7}{*}{Updating}
            & \textbf{Locating and Updating Emergeng Bricks}: Knowledge Locating~\citep{DBLP:conf/emnlp/GevaSBL21,DBLP:conf/acl/DaiDHSCW22,DBLP:conf/icml/SundararajanTY17,DBLP:conf/nips/MengBAB22,DBLP:journals/corr/abs-2301-04213}, Knowledge Updating~\citep{DBLP:journals/corr/abs-2012-00363,DBLP:conf/nips/MengBAB22,DBLP:conf/iclr/MengSABB23,DBLP:conf/acl/OnoeZPDC23,DBLP:journals/corr/abs-2306-09306,DBLP:conf/emnlp/CaoAT21,DBLP:conf/iclr/MitchellLBFM22} \\
           && \textbf{Injecting New Customized Bricks}: Plug-and-Play Knowledge Injection~\citep{DBLP:journals/corr/abs-2210-04726,DBLP:conf/iclr/HuangSZZR023,DBLP:conf/icml/MitchellLBMF22,hernandez2023inspecting,DBLP:journals/corr/abs-2308-10248,DBLP:journals/corr/abs-2310-01405} \\
        \cmidrule{2-3}
        & \multirow{7}{*}{Growing}
            & \textbf{Growing for Pre-training}: Progressive Dense Parameter Growing~\citep{gong2019efficient,gu2021transformer,DBLP:conf/acl/QinZLL0SZ22,DBLP:conf/iclr/WangPHGKFCWK23,wu2024llama}, Progressive Sparse Expert Growing~\citep{DBLP:journals/corr/abs-2208-03306,DBLP:conf/iclr/KomatsuzakiPLRM23,DBLP:journals/corr/abs-2306-04640,wang2024self} \\
           && \textbf{Growing for Post-training}: Multitask Continual Learning~\citep{DBLP:conf/acl/MahabadiR0H20,DBLP:conf/cvpr/0002ZL0SRSPDP22,DBLP:conf/eccv/0002ZESZLRSPDP22,DBLP:conf/iclr/RazdaibiedinaMH23,DBLP:conf/emnlp/MadottoLZMCLYCF21,DBLP:journals/corr/abs-2309-14763} \\
    \bottomrule
    \end{tabular}
    
    \label{tab:refsum}
\end{table*}

\textbf{\textit{Problem 2: how can we leverage existing bricks to build configurable foundation models for the ever-increasing complex requirements of real-world tasks?} ($\S\ \ref{operations}$)}

Bricks consist of a collection of parameters with specialized functionalities. In real-world tasks, singular knowledge and capabilities frequently fall short of meeting task requirements, implying that we need to combine multiple bricks to understand instructions and accomplish specific tasks effectively. 
In this paper, we summarize four primitive operations on bricks and argue that \textit{through the composition of these primitive operations, configurable foundation models can be built based on bricks to fulfill complex task instructions efficiently.} The operations are summarized as follows:
\begin{itemize}[leftmargin=*]
\item 
\textbf{Routing and Retrieving}. The routing and retrieving operation involves the dynamic selection of specific bricks from a brick repository based on instructions. This operation acts as a dynamic brick gatekeeper and allows the configurable foundation model to adapt its brick composition to the instruction at hand. ($\S\ \ref{router}$)

\item 
\textbf{Combining}. The combining operation involves the synergistic integration of multiple bricks, facilitating collaborative and comprehensive processing. This can be achieved by directly merging isomorphic bricks to enhance abilities, or by simultaneously inserting multiple bricks to create composite skills. This operation empowers the model to harness the collective capabilities of various bricks, facilitating the creation of informative responses. ($\S\ \ref{combination}$)

\item 
\textbf{Updating}. The updating operation involves the refinement and adaptation of bricks over time, based on new knowledge and feedback. This operation enables bricks to be fine-tuned or adjusted to improve their performance continually. It also empowers foundation models to adapt and remain pertinent in dynamic real-world scenarios. ($\S\ \ref{updating}$)

\item 
\textbf{Growing}. The growing operation pertains to the expansion of the brick repository itself. New bricks with specialized functionalities can be added to the repository to address emerging requirements. By incorporating new modules, configurable models can keep up with the increasing complexity of real-world applications and offer effective solutions to a broader range of challenges. ($\S\ \ref{growing}$)
\end{itemize}


Besides the above two important problems, in this paper, we conduct empirical analysis on widely-used LLMs to investigate whether these existing well-trained LLMs exhibit functional partitioning similar to the human brain. The experimental results from two models, Llama-3-8B-Instruct and Mistral-7B-Instruct-v0.3, demonstrate that: (1)~Neuron activation is sparse, meaning that processing each instruction requires only a small subset of neurons. (2)~Neurons are specialized for specific functionalities, with the removal of these neurons having minimal impact on other capabilities. (3)~There is evidence of neuronal partitioning, indicating that different capabilities require distinct sets of neurons.

In the end, we discuss the future research directions for the application of configurable foundation models ($\S\ \ref{discussion}$), including: 

\begin{itemize}[leftmargin=*]
\item 
\textbf{Analyzing the correlation between emergent and customized bricks}: Here, we focus on delineating roles between emergent and customized bricks, as well as identifying and handling knowledge conflicts and redundancies arising from their interaction. ($\S\ \ref{correlation}$)

\item 
\textbf{Unifying the protocol to construct bricks}: We engage in a discourse on a novel paradigm for developing foundation models, wherein the shift moves from training a whole model to training individual bricks. This envisioned paradigm entails a shared core foundation model for open-source communities, enabling individuals to develop their bricks based on a unified protocol and openly share bricks for collective utilization. ($\S\ \ref{protocol}$)

\item 
\textbf{Evaluating configurable models}: This facet centers on evaluating the foundation models from the perspective of bricks and discussing the evaluation metrics for configurable bricks. ($\S\ \ref{safety}$)

\item 
\textbf{Implementing the framework for efficient computing}: Our deliberation encompasses the foundational computational operators of configurable models, characterized by sparsity and computational decoupling. Additionally, we delve into the prospects of distributed center-edge computing frameworks for configurable foundation models. ($\S\ \ref{compute-framework}$)

\item 
\textbf{Combining multiple model-level bricks for composite capability}: 
In the rapidly evolving AI community, a vast array of large pre-trained models has been open-sourced, which can serve as model-level bricks for completing complex instructions. We discuss the potential and challenges for scalable multi-model cooperation systems. ($\S \ref{multi-model}$)
\end{itemize}

In summary, we present the concepts coined in this paper in Table~\ref{tab:concept} and present existing representative efforts for configurable foundation models in Table~\ref{tab:refsum}.
We aspire for our paper to serve as an inspiration for future researchers, driving forward the progress of efficient and scalable foundation models.

\section{Configurable Foundation Models}
\label{framework}
In this section, we elaborate on the general framework for configurable foundation models, consisting of various bricks. These bricks encompass both the emergent bricks from the pre-training process and customized bricks from post-processing to enhance LLMs. Specifically, we first present that the pre-trained LLMs naturally possess the property of modularity and can be split into bricks with pre-defined structures or self-organized functional neuron clusters ($\S\ \ref{emergent}$). Then, in the pursuit of advancing LLMs, it is promising to parameterize the external knowledge and capacities into neural bricks, which can be inserted into LLMs in a plug-and-play manner ($\S\ \ref{plugin}$). 
Subsequently, we discuss how to select the granularity of bricks to trade off efficiency and effectiveness ($\S\ \ref{granularity}$). Lastly, we present five benefits to constructing LLMs with configurable bricks, including high efficiency, reusability, traceability, sustainability, and distributed computation~($\S\ \ref{benefit}$).

\subsection{Emergent Bricks}
\label{emergent}
The emergent property of modularity has been observed in the pre-training process of language models~\citep{DBLP:conf/emnlp/WangWZ0LL22,DBLP:conf/acl/ZhangZLXW00XS023}, which indicates that a subset of the parameters can function properly as the entire model for specific instructions. 
Such property makes it possible to break down the gigantic LLMs, including both dense models~\citep{DBLP:conf/nips/BrownMRSKDNSSAA20, DBLP:journals/corr/abs-2302-13971} and sparse models~\citep{DBLP:conf/iclr/LepikhinLXCFHKS21, DBLP:journals/jmlr/FedusZS22}, into tiny modules.
With the breakdown, the aforementioned issues of efficiency and scalability can be tackled via module dropping~\citep{DBLP:conf/iclr/FanGJ20, DBLP:conf/aaai/ZengHVPC23, DBLP:conf/icml/LiuWDZY0S0TRC23}, subnetwork extraction~\citep{DBLP:conf/iclr/FrankleC19,wang-etal-2020-structured,xia-etal-2022-structured}, and recombination~\citep{DBLP:conf/naacl/GururanganLHSZ22,DBLP:conf/acl/ZhangL00S022}.
We term these modules directly broken down from the pre-trained models as emergent bricks, which acquire certain capabilities of the entire model from the pre-training process. In this subsection, we summarize the potential inspirations for emergent bricks and introduce two different categories of emergent bricks, including bricks with human-defined and self-organized structures.
The discussion may boost our understanding of the working mechanism inside the LLMs and help us better configure the LLMs with various internal modules.

\subsubsection{Observations on Parameter Differentiation}

LLMs tend to be over-parameterized when performing some specific tasks~\citep{DBLP:conf/icml/HoulsbyGJMLGAG19,DBLP:journals/natmi/DingQYWYSHCCCYZWLZCLTLS23}, which indicates that there exists a sub-module functioning nearly the same as the entire model with the rest parameters being redundant. This over-parameterization phenomenon leads to two general questions: (1) \textit{``Which part of the model is actually functioning?''} (2) \textit{``What kind of ability does it have?''}. 
In this subsection, we discuss existing observations about the functional specialization of internal parameters in LLMs, that is, each parameter is only responsible for a specific function.


\textbf{Activation Sparsity} 
Inspired by the sparsity in human brains~\citep{olshausen1996emergence,kerr2005imaging,poo2009odor,barth2012experimental} that only a small portion of the neurons activate at each time, special architectures such as sparsely-activated Mixture-of-Experts (MoE) are introduced into Transformers to enforce activation sparsity and thus improve model efficiency~\citep{DBLP:conf/icml/DuHDTLXKZYFZFBZ22,DBLP:conf/icml/RajbhandariLYZA22,DBLP:conf/nips/JaszczurCMKGMK21,DBLP:journals/jmlr/FedusZS22,DBLP:journals/corr/abs-2209-01667}. 
Different from the sparsity of expert activation in MoE, researchers also explore the activation sparsity of model neurons, which is in finer granularity.
Specifically, the neuron ``activation'' refers to the intermediate output of the fully connected layer after the non-linear activation function, and ``sparsity'' indicates that only a few entries of the activation values are nonzero for each given input.
\citet{DBLP:conf/acl/ZhangL00S022} inspect the computational pattern of pre-trained Transformers and find that the activation sparsity naturally exists in pre-trained dense Transformers. Specifically, they delve into the feed-forward networks (FFNs), which constitute two-thirds of the Transformer model parameters, and find the emergence of sparse activation (e.g., only around $5\%$ of the neurons are with nonzero activation values for $90\%$ of the input for a fine-tuned T5-Large model~\citep{DBLP:journals/jmlr/RaffelSRLNMZLL20}). \citet{DBLP:conf/iclr/LiYBLRRYCYGK23} comprehensively investigate sparse activation in Transformers and conclude that it is a ubiquitous phenomenon that emerges for both natural language and vision models, on both training and evaluation data, on datasets of varying scale, on Transformers of varying configurations, and across all layers of a Transformer. 
Although the above works focus on the sparsity within ReLU-based models, an increasing number of modern LLMs have been trained with non-ReLU activations, and it is more tricky to explore activation sparsity for them since there are typically many near-zero but nonzero small activation values. However, recent research shows that these non-ReLU models may be converted to ReLU versions without major performance degradation by fine-tuning with ReLU activations~\citep{DBLP:conf/acl/ZhangL00S022,mirzadeh2023relu}, making activation sparsity pragmatic. 
Moreover, \citet{zhang2024relu} shows that for non-ReLU models, there are still some neurons whose outputs are close to zero and can be discarded without performance degradation, which also indicates the existence of activation sparsity in non-ReLU models.
In summary, activation sparsity refers to the phenomenon that only a small portion of weights play a role for each input, including both expert activation sparsity in MoE and neuron activation sparsity in dense models, and it is different from the sparsity in the weight matrix leading to pruning~\citep{DBLP:conf/iclr/FrankleC19,wang-etal-2020-structured,xia-etal-2022-structured}. Besides, activation sparsity can greatly accelerate the inference process by involving only parts of the parameters for computation~\citep{song2023powerinfer,xue2024powerinfer}.


\begin{figure}
    \centering
    \includegraphics[width=0.98\textwidth]{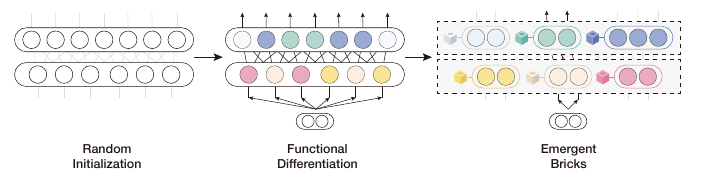}
    \caption{The illustration for emergent bricks. In the randomly initialized, functional differentiation emerges following the pre-training phase. Neurons with similar functionalities can be aggregated to form small functional bricks. Here, the model is divided into two human-defined layer bricks, of which each is further subdivided into several self-organized expert bricks.}
    \label{fig:emergent-bricks}
    \vspace{-1em}
\end{figure}
\textbf{Function Localization} 
In addition to activation sparsity, neurons in the human brain exhibit a modular characteristic: neurons with similar functions tend to cluster together to form specific functional partition~\citep{bullmore2009complex, meunier2010modular}. Similarly, it is widely reported that substantial functions are specifically localized in a small number of parameters within pre-trained models, i.e., ``neurons'' or ``circuits'': 
(1) \textit{Knowledge Neuron}. \citet{DBLP:conf/acl/DaiDHSCW22} and \citet{DBLP:conf/nips/MengBAB22} find that factual knowledge tuplets are stored in neurons of FFNs, and manipulating the activations or weights of these ``knowledge neurons'' can effectively edit the knowledge-related predictions of LLMs. 
(2) \textit{Skill Neuron}. Some researchers dive into finding skill neurons, of which the activations are highly predictive of the task labels. These skill neurons are task-specific and perturbations to skill neurons can drastically impair the performance of corresponding tasks, implying that the task skills are surely localized in these neurons~\citep{DBLP:conf/emnlp/WangWZ0LL22,DBLP:journals/corr/abs-2309-12263,DBLP:conf/icml/PanigrahiSZA23}. 
(3) \textit{Linguistic Neuron}. \citet{DBLP:journals/corr/abs-2305-01610} and \citet{DBLP:journals/corr/abs-2309-04827} study the linguistic features encoded in neurons and observe that neuron activations have correlations with a wide range of features like n-grams and positions. \citet{zhao2023unveiling} and \citet{tang2024language} discover language regions of LLMs that are specialized in multilingual text processing. 
Inspired by these observations, \citet{DBLP:conf/acl/ZhangZLXW00XS023} expand the neuron analysis to include MoE (Mixture of Experts) experts, which represent clusters of neurons. It demonstrates that these sparsely-activated experts are specialized in different functions including knowledge storing, task skills, and semantic understanding.

\subsubsection{Human-Defined Emergent Bricks} 

Neural networks are usually constructed by stacking multiple human-designed network modules at different granularity levels. For example, the original Transformer~\citep{DBLP:conf/nips/VaswaniSPUJGKP17} model consists of multiple identical blocks, within which there are multi-head attention (MHA) layers and FFNs. Simultaneously, each of the MHA layers is a combination of multiple attention heads and each FFN can be viewed as a collection of single artificial neurons that can be further regrouped based on their activations. 
Here, the structures and connections of these stacked modules, including neurons, attention heads, and whole layers, are explicitly designed by humans. Thus, we term these bricks as human-defined emergent bricks.

After defining the structures of neural models, the next pivotal step to empower each brick with problem-solving capabilities is training, to which there are generally two main approaches: 

\textbf{End-to-end Training} End-to-end training of the entire network, which is the most common practice in the deep learning community. In this case, the skills or abilities of each module are not explicitly defined. Existing intriguing observations find that these human-defined bricks can gradually become functionally specialized during the end-to-end training process of the full model~\citep{DBLP:conf/acl/ZhangZLXW00XS023}. For a specific task or input, only a subset of these emergent bricks are functional or informative with others being redundant, and there can be multiple granularity levels of the potential emergent bricks. 
Here we give three examples from top to bottom.
\citet{DBLP:conf/iclr/FanGJ20} show that it is possible to reduce the model computational costs by selecting only some of the layers from a pre-trained model. \citet{DBLP:conf/nips/MichelLN19} find that simply using one attention head can achieve comparable performance to the full model on certain tasks. \citet{DBLP:conf/naacl/ZuoZLHZC22} identify the important neurons in the FFNs for a specific task and then reorganize the sub-network based on the importance of different neurons to achieve better efficiency. 
The intrinsic modular characteristic of these human-defined bricks ensures the effectiveness of the parameter pruning methods. 

\textbf{Modular Training} Modular training conducts training for different modules separately, with their functionalities predefined in advance. \citet{DBLP:conf/cvpr/AndreasRDK16} first propose the neural module networks for visual question answering, where multiple question-specific modular networks are dynamically initiated according to different reusable functional components (e.g., for recognizing dogs, and classifying colors). 
In this case, each of the reusable components represents a human-defined emergent brick that possesses predefined functionality. Such training paradigm also appeals to Transformer-based models, characterized by the MoE models~\citep{DBLP:conf/naacl/GururanganLHSZ22,zhang2022skillnet}, where each expert brick is responsible for one domain or one task. 
Based on these human-defined emergent bricks with predefined functionalities, it is cost-efficient to construct a new task-specific model by intentionally combining part of the bricks within a general-purpose pre-trained model.

Generally, human-defined emergent bricks are commonly observed and hierarchically structured, which naturally expedites the research on various brick configurations. However, there still exist inescapable obstacles to configuring foundation models with human-defined bricks: (1) The functionalities of human-defined bricks acquired by end-to-end training are hard to interpret and localize, preventing them from being clearly and effectively utilized; (2) Modular training of human-defined bricks requires delicate design of each single brick, such as the structure and scale of the modules, to ensure that each brick can effectively attain the functionality predefined by humans in advance. 

\subsubsection{Self-Organized Emergent Bricks}
Language models acquire various capabilities from the pre-training process, which are further stored as parametric knowledge within the model parameters. Observations from the sparse activation and function localization phenomenon imply that the internal parametric knowledge or capabilities are not universally distributed among the entire model, but instead are stored in a centralized manner. However, such centralized distribution of the parametric knowledge does not strictly follow the human-defined module structure, as most modern LLMs adopt the end-to-end training paradigm which intrinsically does not constrain the learning objective of each module explicitly. In this case, there must exist an implicit structure of parametric knowledge distribution that differs from the human-defined module structure and the universal distribution over the entire model. Such an implicit structure is naturally formed during the training process, where different parts of the model interact and collaborate without explicit instructions by humans. Therefore, though language models are built upon human-defined bricks, dependencies and connections between bricks also emerge during the training process, resulting in self-organized emergent bricks.

Distinct from any individual human-defined brick, the concept of self-organized bricks emerges from the interaction between multiple human-defined bricks. There have already been preliminary explorations of self-organized bricks in the literature shedding light on their characteristics and implications. For example, \citet{DBLP:conf/acl/ZhangL00S022} find that some small proportion of the neurons within the Transformer feed-forward networks tend to activate together while the rest of the neurons are inactivated, indicating that such subset of neurons are self-organized to function properly after pre-training and thus give rise to a new form of emergent bricks that is different from those pre-defined by humans. Following this finding, \citet{DBLP:journals/corr/abs-2310-04361} enforce the activation sparsity and only adopt the self-organized bricks for improvement in inference efficiency. Furthermore, \citet{DBLP:conf/acl/ZhangZLXW00XS023} demonstrate that these self-organized groups of neurons possess high productivity for specific functions and any perturbations to them can lead to drastic degradation of the performance in the corresponding function, implying that such self-organized neuron clusters can serve as emergent bricks and are functionally specialized. \citet{DBLP:conf/icml/LiuWDZY0S0TRC23} dive deeper into the concept of self-organized bricks by exploring both Transformer feed-forward networks and muti-head attention layers. Specifically, based on the discovery that the residual connections \citep{DBLP:conf/cvpr/HeZRS16} in LLMs make token embeddings barely change across different layers, they envision input-dependent subsets of neurons in feed-forward networks and attention heads to yield performance comparable to employing the entire model. Moreover, \citet{mirzadeh2023relu} introduce ReLU non-linear activation functions into the layer normalization of Transformers, which further enhances the collaboration among human-defined bricks and leads to more compact self-organized bricks. These works suggest that the functionality of the monolithic LLMs relies on the interaction and collaboration between multiple human-defined bricks at different granularities, and these newly clustered parameters form the self-organized bricks that are specialized in certain functions. 

Introducing self-organized bricks is beneficial to improving both the efficiency and interpretability of language models. First, it is possible to decompose a task-specific sub-model with minimal cost from an existing LLM based on its self-organized bricks, which are more compact compared with its human-defined bricks. For example, we can replace the conventional layer-wise pruning with a more concise selection of functional neuron subsets in the feed-forward networks to improve efficiency. Second, we can align the functionality to the self-organized bricks more flexibly than the human-defined bricks without the explicit structure constraint. Hence, the inner working mechanism of the model can be better interpreted by analyzing the status of the various self-organized bricks.

Though preliminary progress has been made in self-organized emergent bricks, there are still several unexplored aspects that demand further attention from the community: (1) \textbf{Inspecting cross-layer organization}: Currently investigated self-organized emergent bricks are relatively flat and are mostly constructed in parallel within homogeneous layers, whereas they can be organized across different layers. For example, the sparse activation of Transformer feed-forward layers and multi-head attention layers, as observed in~\citet{DBLP:conf/icml/LiuWDZY0S0TRC23}, could correlate to each other with chained dependencies. (2) \textbf{Enhancing training strategy}: Though the existence of self-organized emergent bricks can be revealed in the models trained in an end-to-end manner, models with explicit modular structure still struggle to learn the modular data distribution with the conventional end-to-end training algorithms~\citep{DBLP:conf/nips/MittalBL22}. Enhanced training strategies should be proposed to explicitly encourage the emergence of self-organized emergent bricks. (3) \textbf{Guiding network design}: Scrutinizing self-organized emergent bricks can provide valuable guidance for designing networks with improved efficiency and interpretability in the future. For example, it is possible to identify the minimum combination of bricks that are necessary for a specific task, which can be further employed as a reference for exploring the scaling law and emergent abilities of existing LLMs.

\subsection{Customized Bricks}
\label{plugin}

A monolithic LLM can be split into several emergent bricks, even when LLM layers are trained end-to-end with fully connected neurons. 
As LLM parameters continue to scale, the number of emergent bricks acquired during pre-training also increases, which enables satisfactory downstream performance. However, as the world continually changes, the capacities and knowledge that models need to master are constantly being updated and expanded. For example, world knowledge contained in widely-used Wikipedia is edited and updated daily~\citep{DBLP:conf/emnlp/CaoAT21,DBLP:conf/nips/MengBAB22}; new academic papers published on scholarly websites continuously advance domain knowledge~\citep{DBLP:conf/naacl/GururanganLHSZ22,DBLP:conf/naacl/JinZZ00WA022}; and novel tasks emerging in different application scenarios also demand ever-growing task capacities from LLMs~\citep{DBLP:journals/corr/abs-2302-00487}. For ease of introduction, we use the term, knowledge, to refer to both knowledge and capacities, which can be regarded as a type of abstract knowledge.

Training the whole model to incorporate new knowledge is computation-intensive and requires massive storage space. To address this issue, many efforts have been devoted into parameterizing the external knowledge as customized bricks, which can be injected into LLMs for performance promotion~\citep{DBLP:conf/iclr/DathathriMLHFMY20,DBLP:conf/emnlp/LauscherLG21,DBLP:conf/acl/ZhangZLWYXHLLSZ23,DBLP:conf/acl/XiaoZHCLLLLCS23,DBLP:journals/natmi/DingQYWYSHCCCYZWLZCLTLS23}.
Specifically, customized bricks are usually constructed after pre-training with the original parameters frozen. Customized bricks can serve as an external knowledge bank for LLMs. Given an instruction, we first retrieve bricks with relevant knowledge and then insert them into LLMs for better responses. Different from training LLMs to store knowledge in emergent bricks, customized bricks possess the plug-and-play characteristic for dynamic and reusable knowledge injection. Therefore, customized bricks are usually named ``plugins''. In this subsection, we will first summarize the underlying reasons for the feasibility of customized bricks and introduce several typical customized bricks.

\begin{figure}
    \centering
    \includegraphics[width=\textwidth]{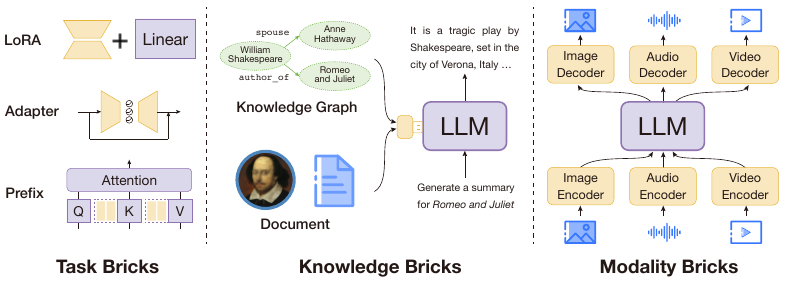}
    \caption{The illustration for three typical customized bricks, including task bricks, knowledge bricks, and modality bricks.}
    \label{fig:plugin}
\end{figure}

\subsubsection{Observations on Intrinsic Dimension of LLMs}
Customized bricks aim to inject external knowledge or new task adaptations into foundation models with tiny neural modules with a few parameters. This raises a natural question: \textit{``Can the external knowledge be represented in limited parameters?''}

\paragraph{Intrinsic Dimensionality} It has been widely recognized that LLMs are highly over-parameterized~\citep{DBLP:conf/iclr/FrankleC19,DBLP:conf/acl/LiangZCJLHZC20,DBLP:conf/emnlp/PrasannaRR20}, which is also the case of almost all the modern neural network models. This brings a natural question of what is the minimal number of parameters needed to describe the training of these models. \citet{DBLP:conf/iclr/LiFLY18} introduce the concept of ``Intrinsic Dimension'' of an objective landscape. For a given training objective landscape, its intrinsic dimension is defined as the minimal number of free variables required to well define the optimization problem, i.e., the minimal possible dimension to reparameterize the objective into. \citet{DBLP:conf/iclr/LiFLY18} propose to estimate the intrinsic dimension by randomly projecting the parameters of the original model into a low-dimensional subspace and observing if the random subspace contains a good enough solution for the training objective. If there is a solution, the dimension of the current subspace serves as an upper bound of the intrinsic dimension. With this method, \cite{DBLP:conf/acl/AghajanyanGZ20} examine the intrinsic dimension of pre-trained language models and find that the fine-tuning of PLMs has a very low intrinsic dimension ($\sim 200$ for RoBERTa) and pre-training implicitly minimizes the intrinsic dimension. It has also been observed that larger models tend to have lower intrinsic dimensions. \citet{DBLP:journals/corr/abs-2110-07867} further find that the PLM adaptations to many different tasks not only commonly have low intrinsic dimension but the many tasks can also be reparameterized into a shared universal low-dimensional subspace, which partially explains the prevalent effectiveness of foundation models and the transferability of parameter-efficient tuning~\citep{DBLP:conf/naacl/SuWQCLWWLLL0SZ22}. \citet{DBLP:conf/acl/ZhangLS23} also find that fine-tuning can be performed within a low-dimensional subspace and some outlier dimensions play an important role. The low intrinsic dimensionality and the universal existence of intrinsic subspace make us believe that adding new abilities and knowledge into foundation models with relatively small-scale customized bricks is possible.

\subsubsection{Typical Customized Bricks}
\label{subsubsec:typical-customized-bricks}

Customized bricks have emerged as a significant avenue for enhancing LLMs during their post-processing phase, which involves the insertion of tiny bricks after the pre-training or fine-tuning procedures. This approach aims at efficiently enhancing the LLM's customized capabilities. As shown in Figure~\ref{fig:plugin}, we categorize widely used bricks into three types based on their capabilities: task bricks, knowledge bricks, and modality bricks.


\subsubsection{Task Bricks}
Task bricks, also known as parameter-efficient tuning or delta tuning~\citep{DBLP:conf/iclr/HeZMBN22,DBLP:journals/natmi/DingQYWYSHCCCYZWLZCLTLS23}, are widely explored as a substitute for full-parameter fine-tuning, which usually requires substantial computational and storage costs. Task bricks adopt task adaptation by tuning only a small portion of parameters. 
As the number of model parameters grows, the performance gap between task bricks and full-parameter fine-tuning narrows. Consequently, in the realm of LLMs, employing task bricks for adaptation has become a widely accepted paradigm.
Following \cite{DBLP:journals/natmi/DingQYWYSHCCCYZWLZCLTLS23}, we divide task bricks into three types according to the operations on the tunable parameters. Besides, recent efforts demonstrate that the task bricks can also be obtained by extracting task vectors from the internal representations without tuning, and we term these efforts as training-free task bricks. 

\textbf{Addition-based Bricks} Addition-based bricks introduce extra parameters into the LLM for fine-tuning. Within this category, the most extensively studied methods are adapter tuning~\citep{DBLP:conf/icml/HoulsbyGJMLGAG19} and prompt tuning~\citep{DBLP:journals/csur/LiuYFJHN23}. The fundamental structure of adapter tuning consists of two linear layers with a notably low intermediate dimension, which enables efficient computation and storage. This layer can be inserted into the standard transformer architecture, for instance, following the self-attention layers or the feed-forward network layers, to facilitate task adaptation. Different from adapter layers to modify the model architectures, prompt tuning conducts task adaptation by inserting token embeddings in the input layers~\citep{DBLP:journals/csur/LiuYFJHN23,DBLP:conf/acl/DingHZCLZS22}. The early researches prepend hard discrete tokens to the inputs, which aims to bridge the gap between pre-training and fine-tuning via formalizing all NLP tasks into sequence generation problem~\citep{DBLP:conf/nips/BrownMRSKDNSSAA20,DBLP:conf/acl/GaoFC20,DBLP:conf/eacl/SchickS21}. Then to make prompt tunable, soft prompt is proposed to prepend randomly initialized continuous embeddings to inputs and optimize task objectives via gradient descent~\citep{DBLP:conf/emnlp/LesterAC21}. Prompt tuning is a human interaction-friendly algorithm as the users can drive LLMs to accomplish various tasks by utilizing different prompts eliminating the need to modify the model's architecture. 


\textbf{Specification-based Bricks} 
Specification-based bricks specify some existing parameters in LLMs to be tunable and do not introduce additional parameters. BitFit~\citep{DBLP:conf/acl/ZakenGR22} presents that only optimizing the bias vectors inside the linear projections can achieve satisfactory performance. \cite{DBLP:conf/acl/CuiLDHLS23} attempt to only tune the output layer to protect data privacy and enable an efficient API-based tuning framework. Besides manually specifying which parameters are adjustable, \cite{DBLP:conf/acl/GuoRK20} and \cite{DBLP:conf/emnlp/ZhaoLMJS20} learn a binary mask for the parameters, determining which ones should be optimized for a given task. When the model scales to billions of parameters, the performance gap introduced by the design differences becomes negligible, and even arbitrarily specifying modules to be tunable can also lead to comparable results with full-parameter fine-tuning~\citep{DBLP:journals/corr/abs-2306-02320}.

\textbf{Reparameterization-based Bricks} Reparameterization-based bricks rewrite the computational formula of existing layers into a parameter-efficient form and specify part of the parameters tunable.
The most widely adopted approaches within this category assume that the variations in some parameters during training are low-rank, subsequently optimizing these low-rank variations with tiny parameters. For instance, the intrinsic dimension, as mentioned earlier, maps a low-dimensional vector into the parameter space, allowing the model training process to solely optimize this low-dimensional vector~\citep{DBLP:conf/acl/AghajanyanGZ20,DBLP:conf/acl/ZhangLS23}. Similarly, LoRA models the variation of a particular parameter matrix as the product of two matrices with significantly low intermediate dimensions~\citep{DBLP:conf/iclr/HuSWALWWC22}.

\textbf{Training-free Task Bricks}
In addition to the aforementioned methods that require additional training, many researchers attempt to activate the intrinsic task capabilities of foundation models without any training.
Recent progress shows that the representation space of intermediate layers in LLMs possesses semantically meaningful structures~\citep{DBLP:journals/corr/abs-2310-01405}. It indicates that we can directly control the behaviors of LLMs by operating the intermediate representations. Inspired by these findings, many efforts reveal that the demonstrations in in-context learning can be transformed into a function vector with simple representation arithmetic, and inserting the function vector into intermediate representations of inputs can trigger the foundation model to generate the task predictions~\citep{DBLP:journals/corr/abs-2311-06668,DBLP:journals/corr/abs-2310-15213,DBLP:conf/emnlp/HendelGG23}.

\subsubsection{Knowledge Bricks}
Knowledge bricks aim to supplement LLMs with external knowledge. While it is well-documented that LLMs internalize amounts of world knowledge to facilitate robust language comprehension~\citep{DBLP:conf/emnlp/PetroniRRLBWM19,DBLP:conf/emnlp/RobertsRS20,DBLP:journals/tacl/JiangXAN20}, their finite parameter space inevitably limits the capacity to encapsulate the nearly infinite spectrum of external knowledge. This limitation often manifests itself in the form of ``hallucinations'', where the model generates erroneous information in responses due to a lack of relevant knowledge~\citep{DBLP:conf/acl/MaynezNBM20,DBLP:conf/acl/LinHE22,DBLP:conf/naacl/HonovichAHTKCSS22,DBLP:journals/csur/JiLFYSXIBMF23}. Moreover, the computational overhead of LLMs make them less agile in adapting to the ever-changing landscape of world knowledge~\citep{DBLP:conf/iclr/WuCLLQH22,DBLP:journals/corr/abs-2302-00487}. In the following paragraphs, we will present methods for representing knowledge as compact neural bricks and discuss the advantages of knowledge bricks compared to traditional knowledge injection methods.
Based on the knowledge type, we can divide the knowledge bricks into structured knowledge graph (KG) bricks and unstructured text bricks. 

\textbf{Structured KG Bricks} Leveraging structured KGs to enhance pre-trained language models has long been a pivotal direction in NLP~\citep{DBLP:conf/acl/ZhangHLJSL19,DBLP:journals/cacm/HanZL21}. Compared to traditional models that incorporate knowledge during the pre-training phase, the construction of tiny structured KG bricks is cost-efficient, with stronger scalability to build bricks across diverse knowledge types. The core of structured KG bricks lies in pluggable knowledge representations, which can be injected into LLMs to provide external knowledge. One main line of research attempts to concatenate informative entity representations with original input embedding sequences for knowledge fusion. To this end, \cite{DBLP:conf/acl/YeL0S022} average the output vectors of masked entity’s occurrences as pluggable representations. \cite{DBLP:conf/emnlp/PornerWS20} and \cite{DBLP:conf/acl/ZhangZLWYXHLLSZ23} learn a neural projection to bridge the gap between KG embeddings~\citep{DBLP:conf/nips/BordesUGWY13,DBLP:conf/conll/YamadaS0T16} and token embeddings in pre-trained models. 
Besides, different from incorporating knowledge features, K-Adapter adopts knowledge-specific objectives, such as entity relation extraction for KG and dependency prediction for linguistic trees, to optimize adapters for knowledge-enriched representations from originally hidden vectors~\citep{DBLP:conf/acl/WangTDWHJCJZ21}.


\textbf{Unstructured Text Bricks}
Unstructured text is the primary medium for humans to record and store world knowledge.
An increasing number of researchers are exploring ways to augment LLMs with an external retrievier, where the relevant textual knowledge is concatenated with input instructions~\citep{DBLP:conf/icml/GuuLTPC20,DBLP:conf/nips/LewisPPPKGKLYR020,DBLP:journals/corr/abs-2112-09332,DBLP:journals/corr/abs-2302-07842}. However, these approaches often suffer from poor reusability of encoded knowledge, requiring redundant knowledge re-encoding for different instructions. 
Therefore, constructing cross-task reusable, plug-and-play textual knowledge bricks presents an efficient method. \cite{DBLP:conf/eacl/SaadFalconSSDCD23} adopt the activations of long documents as reusable representations. During downstream inference, the activations are directly fed into the top layers of pre-trained models to reduce computational overhead. \cite{DBLP:conf/acl/XiaoZHCLLLLCS23} represent documents as prefix tokens and conduct self-supervised training to enable document bricks suitable for both inference and fine-tuning.

\subsubsection{Modality Bricks}
Multimodal large language models (MLLM), which utilize LLMs as the brain for reasoning and are capable of processing various perceptual signals such as images and speech, have become pivotal in the pursuit of artificial general intelligence. With the continuous growth in LLM training data and parameter scale, LLMs have exhibited numerous surprising emergent abilities, including instruction-following, in-context learning, and chain-of-thoughts~\citep{DBLP:journals/tmlr/WeiTBRZBYBZMCHVLDF22,DBLP:conf/nips/Wei0SBIXCLZ22}. To leverage these remarkable abilities in multimodal tasks and scenarios, many researchers have shifted their focus towards the training of MLLM~\citep{DBLP:journals/corr/abs-2306-13549}. However, building an MLLM from scratch necessitates substantial computation and multimodally aligned data pairs. As a result, much of the current work treats pre-trained models from other modalities as bricks for LLMs, effectively leveraging them to transform multi-modal signals into features that the LLMs can readily process and understand. Based on the types of interface features communicated between models, these models can be categorized as bricks with textual interface and continuous interface.

\textbf{Bricks with Textual Interface} 
This line of work initially converts multi-modal data into text, which is then combined with textual instructions and fed into the LLMs. For example, \cite{DBLP:journals/corr/abs-2305-06355} adopts open-source video caption and detection models to convert videos into textual descriptions, which are then fed into LLM to generate responses. Recent popular tool-augmented LLMs treat models for other modalities as APIs. When given the functional descriptions and input-output formats of models, LLMs decompose the input instruction into multiple sub-tasks, where various models are involved to solve one by one~\citep{DBLP:journals/corr/abs-2303-04671,DBLP:journals/corr/abs-2303-17580,DBLP:journals/corr/abs-2303-11381}.
These approaches require no additional training and do not necessitate access to the model's parameters, making them particularly suitable for API-based models such as ChatGPT and GPT-4~\citep{DBLP:journals/corr/abs-2303-08774}. However, converting other modalities into text often results in inevitable information loss, which can considerably impact the model's performance.

\textbf{Bricks with Continuous Interface} 
To eliminate the information loss, many researchers attempt to construct a learnable continuous interface between LLMs and other pre-trained models. As the models are trained separately on single-modality data, the representation spaces between these models are quite different, which poses challenges for the learnable interface to translate visual and audio inputs as LLM continuous prompts. As for the model architecture, a widely-used method is adopting attention mechanism to extract important information with several learnable query vectors~\citep{DBLP:conf/nips/AlayracDLMBHLMM22,DBLP:conf/icml/0008LSH23,DBLP:journals/corr/abs-2305-04160}. Further investigation indicates that a simple multi-layer perceptron is powerful for modalities bridge~\citep{DBLP:journals/corr/abs-2304-08485,DBLP:journals/corr/abs-2304-10592,DBLP:journals/corr/abs-2305-15023,DBLP:journals/corr/abs-2309-05519}. Different from bricks with textual interface, learnable continuous interface relies heavily on multi-modality aligned data~\citep{DBLP:journals/corr/abs-2305-11540,DBLP:journals/corr/abs-2305-16103}. Besides, due to the limited representation capacities of continuous prompts, the learnable interface sometimes suffers from fine-grained information loss.


Besides the three types of customized bricks mentioned above, many researchers devote efforts to developing plugins with diverse functionalities. These include plugins that enable models to manipulate external tools~\citep{DBLP:journals/corr/abs-2301-12652,DBLP:conf/acl/YuXY023,DBLP:journals/corr/abs-2305-11554}, debias the model response~\citep{DBLP:conf/iclr/DathathriMLHFMY20}, reduce the computational costs~\citep{DBLP:conf/emnlp/XiaoLZZHLZXL0Z23}, and transform the style of generated text~\citep{DBLP:conf/emnlp/PascualEMCW21,DBLP:conf/iclr/DathathriMLHFMY20}. The practice of constructing tiny customized bricks - plugins for LLMs to supplement their functionalities and knowledge has become a widely accepted paradigm. Despite this success, the plugin learning for LLMs is still faced with the following challenges:
(1) \textbf{Combining multiple plugins}: In real-world scenarios, it often becomes necessary to combine multiple plugins to execute complex commands. However, since different types of plugins are trained independently, combining them during the inference stage can lead to out-of-distribution (OOD) problems. Exploring the combination of multiple plugins is therefore crucial to unlocking the full potential of large model plugins.
(2) \textbf{Unified training strategy}: Currently, different training methods, datasets, and insertion points are required for plugins with different capabilities. Discussing the construction of different types of plugins from a unified perspective could greatly benefit the future development of numerous plugins. A standardized approach to training would streamline the process, ensuring consistency and efficiency across different plugin types, which would also benefit compatibility issues.

\subsection{Brick Granularity}
\label{granularity}

As stated previously, the granularity of a brick is highly customizable, ranging from a solitary neuron to a whole pre-trained model.
As the size of bricks increases, their capacity expands correspondingly and the computational resources required also increase. This presents a challenge in selecting the optimal brick size, necessitating a careful balance between efficiency and effectiveness.
In this section, we will first review existing observations on the capability of four different granularity of bricks: the solitary neuron ($\S\ \ref{sec:granularity_neuron}$), the neuron group ($\S\ \ref{sec:granularity_group}$), the layer ($\S\ \ref{sec:granularity_layer}$), and the full model ($\S\ \ref{sec:granularity_model}$). Furthermore, we discuss how to choose the brick granularity properly~($\S\ \ref{sec:granularity_discussion}$).

\subsubsection{Solitary Neuron Granularity}
\label{sec:granularity_neuron}

The neuron, defined as a row or a column in the weight matrix of the linear layer, is often considered the finest functional unit in Transformer-based foundation models~\citep{DBLP:journals/corr/abs-2005-07647,DBLP:conf/acl/ZhangZLXW00XS023}. After being trained properly, they can carry certain skills or knowledge, laying a solid foundation for the complex behavior of the entire deep learning system.

From the perspective of skills, neurons in well-trained neural networks are demonstrated to possess the ability to capture specific input patterns and predictivity for some basic NLP tasks. Some early works find that neurons can learn the position of words~\citep{DBLP:journals/corr/KarpathyJL15} or parts of speech~\citep{DBLP:conf/aaai/DalviDSBBG19} such as nouns, verb forms, articles, numbers, etc. Others prove the specialization of certain neurons in capturing groups of words with similar meanings (e.g., electronic items or legislative terms)~\citep{DBLP:conf/naacl/LiCHJ16,DBLP:journals/coling/KadarCA17,DBLP:conf/iclr/NaCLK19}. Further, recent studies demonstrate the potential of visual model neurons in learning meaningful perceptual concepts such as the tall structures in images~\citep{DBLP:conf/nips/MuA20}. The sensitivity of neurons to various input patterns constitutes their high predictivity for some fundamental NLP tasks, including sentiment analysis, natural language inference, topic classification, etc~\citep{DBLP:conf/emnlp/WangWZ0LL22}.

Moreover, factual knowledge is also an important aspect of information that can be obtained by neurons. An important work in this field is done by~\cite{DBLP:conf/acl/DaiDHSCW22}, who demonstrates the potential storage of factual knowledge in specific neurons (i.e., knowledge neurons). The activation of these knowledge neurons is positively correlated to the expression of the corresponding knowledge triplet, which sheds light on a promising approach to training-free knowledge editing and model manipulation~\citep{DBLP:journals/tacl/SajjadDD22}.

Another interesting fact found in previous work is that the skills or knowledge contained in a solitary neuron can even be non-singular. Polysemous neurons capturing multiple concepts or word senses widely exist in deep neural networks~\citep{DBLP:conf/emnlp/XinLY19,DBLP:journals/corr/abs-2005-07647}. The knowledge neurons were responsible for different factual knowledge are also proven to have intersections \citep{DBLP:conf/acl/DaiDHSCW22}. This observation underscores the potential of exploring smaller units within LLMs for understanding the storage of knowledge and skills.

\subsubsection{Neuron Group Granularity}
\label{sec:granularity_group}

Neuron groups, namely tiny sublayers involving a group of neurons, can often display more complex behaviors than solitary neurons.
As demonstrated in~\cite{DBLP:conf/acl/ZhangZLXW00XS023}, neurons can be emergently clustered into different function groups during the pre-training. Besides, the customized bricks usually consist of tiny sublayers to store certain knowledge and abilities. 

One of the most popular organization forms of neuron groups is Mixture-of-Expert (MoE), where each expert is a specialized neuron group and the MoE output is aggregated from the expert outputs through a routing function. As for Transformers, pre-defined MoE is usually implemented by replacing a single linear layer in the attention module~\citep{DBLP:conf/emnlp/ZhangSHZR022} or the feedforward network~\citep{DBLP:journals/jmlr/FedusZS22} with multiple linear experts. In some previous works~\citep{DBLP:conf/emnlp/ZhangSHZR022,DBLP:journals/corr/abs-2202-08906,DBLP:conf/acl/ZhangZLXW00XS023}, these experts are demonstrated to possess specialized functions at different levels, ranging from simple semantic functions (e.g., word sense classification), knowledge functions (e.g., factual knowledge recognition), to more complex language understanding tasks such as GLUE~\citep{DBLP:conf/iclr/WangSMHLB19}. Other works also provide clues for expert specialization through the analyses on expert activations, expert usage, or ablation studies~\citep{DBLP:journals/corr/abs-2212-08066,DBLP:conf/nips/MustafaRPJH22, zhao2024self,shen2023mixtureofexperts,DBLP:journals/corr/abs-2306-04640}. They also demonstrate that the effectiveness of MoE and expert specialization are consistent in textual, visual, and multimodal models. In addition to the pre-defined MoE structure, we can also probably explore MoE structures inside an already pre-trained model without additional parameters. For instance, \cite{DBLP:conf/acl/ZhangL00S022} constructs experts by splitting the FFN parameters into functional partitions, which reduces the computation significantly without harming the overall performance.

Another line of research focusing on a group of neurons is parameter-efficient tuning. To reduce the huge computation costs of tuning large language modules, PET only updates a small number of neurons (inherently inside the model or additionally introduced) while freezing the remaining parts of the model~\citep{DBLP:journals/natmi/DingQYWYSHCCCYZWLZCLTLS23}. The tunable neuron groups in representative PET methods (e.g., Adapter~\citep{DBLP:conf/icml/HoulsbyGJMLGAG19}, Prefix-Tuning~\citep{DBLP:conf/acl/LiL20}, BitFit~\citep{DBLP:conf/acl/ZakenGR22}, and LoRA~\citep{DBLP:conf/iclr/HuSWALWWC22}) are demonstrated to have high versatility and satisfactory performance on over 100 NLP tasks, from simple text classification to complex conditional generation~\citep{DBLP:journals/natmi/DingQYWYSHCCCYZWLZCLTLS23}. PET neuron groups can also carry external knowledge to empower the frozen language model in a plug-and-play manner~\citep{DBLP:conf/acl/ZhangZLWYXHLLSZ23,DBLP:journals/corr/abs-2307-06029}.

\subsubsection{Layer Granularity}
\label{sec:granularity_layer}

The utilization of stacked layers within deep models has consistently showcased its superior performance across numerous scenarios~\citep{wang-etal-2019-learning-deep,DBLP:conf/naacl/DevlinCLT19}. LLMs typically comprise multiple layers with the same architecture, each possessing unique parameters~\citep{DBLP:journals/corr/abs-2302-13971,DBLP:journals/corr/abs-2307-09288}. {\em Understanding} and {\em manipulating} different layers are both crucial aspects to maximizing the potential of LLMs.

Numerous studies have delved into the examination of the functions of distinct model layers.
For instance, \citet{lin-etal-2019-open} present that the word order information is mostly contained in lower layers, while \citet{hewitt-manning-2019-structural} propose the structural probing framework and find that syntactic knowledge is most prominent in middle layers. \citet{liu-etal-2019-linguistic} also find that final layers are usually more related to specific tasks.
\citet{rogers-etal-2020-primer} provide a relatively thorough survey about the function of each layer in BERT~\citep{DBLP:conf/naacl/DevlinCLT19}. In addition to linguistic functions, \citet{DBLP:conf/emnlp/GevaSBL21} demonstrate that the feed-forward layers in Transformer models~\citep{DBLP:conf/nips/VaswaniSPUJGKP17} can be construed as key-value memories, which inspired a new way to addressing and editing the knowledge stored in language models. 
As models progress from lower to higher layers, the functional scope transitions from local, lexical aspects to global, semantic dimensions. 

Based on the observations, there exist several approaches for the manipulation of inner model layers to improve efficiency, among which a straightforward one is to swap out some layers to accelerate model training or inference. \citet{NEURIPS2020_a1140a3d} present a Switchable-Transformer block, which introduces a gate to determine whether the corresponding layer is disabled or not in the training stage, based on which they further proposed a progressive-layer-dropping approach that can effectively reduce the training cost. Regarding inference, \citet{DBLP:conf/acl/XinTLYL20} introduces DeeBERT, which reduces inference time by bypassing certain upper layers rather than passing through the entire model. Another branch of layer manipulation is knowledge editing, which aims to change the existing knowledge within pre-trained models by modifying some specific layers. For example, \citet{huang-etal-2023-knowledge} proposed to inject multilingual knowledge into the feed-forward layers. Recently, \citet{wang2023easyedit} present EasyEdit, which supports various cutting-edge knowledge editing methods and applies to representative large language models, such as Llama 2~\citep{DBLP:journals/corr/abs-2307-09288}.

\subsubsection{Full Model Granularity}
\label{sec:granularity_model}

Various types of models frequently exhibit distinct advantages and drawbacks. For instance, large models excel in performance, while small models offer higher speed and demand fewer computational resources. Combining existing models is an efficient strategy for harnessing the strengths of individual models. \citet{li-etal-2021-cascadebert-accelerating} propose to dynamically select models of different sizes for input samples with different difficulties. Inspired by the dual-process theory of human cognition, \citet{lin2023swiftsage} propose to employ a small model that performs fast and intuitive thinking and a large model for complex and slow thinking.
In addition to amalgamating models of varying sizes, as discussed in $\S \ref{plugin}$ the integration of models from different modalities offers a practical approach to constructing multimodal models~\citep{DBLP:conf/icml/0008LSH23}. 

The studies mentioned above primarily concentrate on the integration of independent models. However, there are also notable works that involve the extraction of sub-networks from larger models.
\citet{pmlr-v139-zhang21a} explore the out-of-distribution generalization capabilities of sub-networks and find that even in biased models there still exist unbiased sub-networks. \citet{xu-etal-2021-raise} identify a sub-network, which they called the child network, in a pre-trained model and only update the weights of the child network for downstream tasks. \citet{xu-etal-2022-s4} also propose S$^4$-Tuning, a technique that partitions the entire model into sub-networks dedicated to each target language. It exclusively updates the relevant sub-network for tasks in a specific language, thereby enhancing language-specific task performance.

\subsubsection{Discussion}
\label{sec:granularity_discussion}

Based on the above statements, we turn back to the issue of selecting the appropriate granularity according to the required ability. Intuitively, coarser-grained bricks with more parameters are better suited for addressing complex tasks. For instance, solitary neurons can discern specific patterns of low-level linguistic units~\citep{DBLP:journals/corr/KarpathyJL15,DBLP:conf/aaai/DalviDSBBG19,DBLP:conf/naacl/LiCHJ16,DBLP:journals/coling/KadarCA17,DBLP:conf/iclr/NaCLK19}. By contrast, full model bricks are typically associated with higher-level capabilities, encompassing the general understanding of specific modalities~\citep{DBLP:conf/icml/0008LSH23}, language, or textual corpus (e.g., codes). However, there exists evidence supporting the presence of certain abilities shared by multiple granularities. Both solitary neurons and neuron groups have been established as having predictive functions in some language understanding tasks, including sentiment analysis and natural language inference~\citep{DBLP:conf/emnlp/WangWZ0LL22,DBLP:conf/acl/ZhangZLXW00XS023}. Consequently, the alignment of diverse granularity levels with specific abilities remains an open question. Scaling laws, the empirical studies on the relationships between task performance, model scale, dataset size, and computation, may provide insights into the granularity-ability association~\citep{DBLP:journals/corr/abs-2001-08361}.

Another point worth attention is that different levels of brick granularities and abilities have potential inclusive relationships. Specifically, high-level abilities, such as general language understanding, can be decomposed into low-level NLP tasks. Similarly, bricks of coarser granularities can be viewed as a combination of multiple finer-grained ones. Therefore, a possible approach involves structuring foundational models with hierarchical bricks. 
Consequently, when retrieving, constructing, or updating bricks, operations can be efficiently executed at the appropriate granularity, guided by the hierarchical functional partitions.
Nevertheless, some works also demonstrate that the functional bricks in LLMs can appear emergently during the training process. Given the yet-to-be-fully-explored association between granularity and abilities, the manual design of well-suited brick hierarchies or abilities presents a challenge. Thus, it becomes necessary to conduct comparative analyses of the performance and costs for different brick granularities, which we leave for future work.

\subsection{Benefits of Configurable Bricks}
\label{benefit}
As elucidated earlier in this section, we can decompose pre-trained models into emergent bricks and enhance their capabilities by constructing custom bricks, thereby realizing a configurable LLM architecture. In this subsection, we will highlight five advantages of configurable LLMs compared to traditional monolithic LLMs.

\textbf{Efficiency}
The vast number of parameters in language models, encapsulating knowledge and capabilities, are essential for executing a wide range of tasks. However, for a specific task or instruction, we often only need to utilize a subset of these parameters for language comprehension and task inference. This means that, given a particular command, we can dynamically select the relevant bricks associated with the current instruction for computation.  
For instance, early-exiting models treat each layer of the model as a brick. For each instruction, they decide whether to engage subsequent layers for computation based on the predicted confidence~\citep{DBLP:journals/corr/abs-2305-16103,DBLP:journals/corr/abs-2102-04906,DBLP:conf/mobisys/LaskaridisKL21}. Models utilizing a mixture-of-experts (MoE) approach regard each expert network (typically an FFN layer) as a brick. For every token, they select a few experts that match the token's characteristics from a set of expert networks to participate in the computation~\citep{DBLP:journals/corr/abs-2202-08906,DBLP:conf/iclr/LepikhinLXCFHKS21}.
Consequently, even as the number of model parameters grows in response to expanding knowledge and capabilities, the computational demand for a given instruction or task remains relatively low.

\textbf{Reusability}
Configurable LLMs decompose the model into several distinct functional bricks, facilitating the performance for real-world complex requirements through combinations of these bricks. In various configurations, knowledge and capability transfer can be achieved by reusing different bricks. For instance, \cite{DBLP:conf/emnlp/PfeifferVGR20} decomposes multi-lingual task fine-tuning into language bricks and task bricks. Training task bricks on a high-resource language and then combining it with a low-resource language brick enables the task knowledge transfer across different languages. Both \cite{DBLP:conf/acl/ZhangZLWYXHLLSZ23} and \cite{DBLP:conf/acl/XiaoZHCLLLLCS23} model KG and textual knowledge as task-agnostic bricks, allowing for efficient knowledge injection across different tasks without the need for further task-specific adaptations.

\textbf{Traceability} 
Decomposing the black-box LLMs into bricks with interpretable functions allows us to trace the underlying mechanisms behind their superior performance by monitoring the activation of bricks.
As aforementioned, efforts have been made to identify the knowledge~\citep{DBLP:conf/acl/DaiDHSCW22}, concept~\citep{DBLP:journals/corr/abs-2005-07647,DBLP:conf/nips/MuA20}, and task-specific skills~\cite{DBLP:conf/emnlp/WangWZ0LL22} encoded in specialized neurons or expert units. When a particular brick is activated, it indicates that the knowledge contained within the brick has been utilized to generate a response to the given instruction. Such observations provide a fresh perspective on understanding the behavior of LLMs.
After the source tracking, we can predictively control model behavior by manipulating relevant bricks without affecting other parts of the architecture~\citep{DBLP:journals/tacl/SajjadDD22}. 
For example, 
\cite{DBLP:conf/acl/DaiDHSCW22} update or erase learned relational facts by directly modifying the parameters in the corresponding knowledge neuron.
This provides a novel viewpoint for the oversight and alignment of LLMs: instead of focusing on a holistic ethical and safety review of the entire LLMs, the focus can shift to a brick-by-brick examination, allowing for the repair or replacement of bricks that may induce the model to generate unethical responses~\citep{DBLP:conf/emnlp/LauscherLG21,DBLP:conf/emnlp/GevaCWG22}.

\textbf{Sustainability}
Enhancing LLMs with continuous capabilities and knowledge to adapt to the ever-evolving global environment remains a focal point in research. Unlike monolithic LLM, which necessitates the updating of whole parameters and can result in catastrophic forgetting of existing knowledge, configurable LLMs can achieve continual learning through the growth and updating of specific bricks without undermining previously acquired knowledge.
For instance, in the realm of multi-domain pre-training, many scholars focus on the MoE model, leveraging the expansion of experts to assimilate knowledge across an increasing array of domains~\citep{DBLP:journals/corr/abs-2208-03306,DBLP:journals/corr/abs-2306-04640,DBLP:conf/iclr/KomatsuzakiPLRM23}. Furthermore, when knowledge embedded within the LLM requires updating or overwriting, strategically modifying specific emergent bricks~\citep{DBLP:conf/emnlp/GevaCWG22,DBLP:conf/nips/MengBAB22}, or constructing a supplementary customized brick~\citep{DBLP:conf/emnlp/DongDSXSL22,DBLP:conf/emnlp/LauscherLG21}, stands as an efficient solution.

\textbf{Distributed Computation}
Configurable LLMs decompose the monolithic computation into modularized operations. The distributed computation trait makes configurable LLMs more practical to deploy on computational clusters: each machine can be tasked with the computation of a specific brick, exchanging information with others through hidden vectors. This distributed computing trait can harness the full computational capacity of each device, thereby reducing deployment costs. For instance, many researchers propose to distribute different bricks across distinct machines, training them with domain-specific data, and eventually merging all the trained bricks to produce a larger language model with enhanced capabilities~\citep{DBLP:conf/icml/WortsmanIGRLMNF22,DBLP:conf/nips/AlayracDLMBHLMM22,DBLP:journals/corr/abs-2208-03306,DBLP:journals/corr/abs-2307-13269}. Moreover, the nature of distributed computing can serve as a safeguard for model and data privacy. For example, \cite{DBLP:conf/acl/CuiLDHLS23} and \cite{DBLP:conf/usenix/ZhouWZS22} place the main LLM on a central server endowed with substantial computational resources, while the task-specific bricks and output layers reside on the user's machine. This setup allows users to reap the benefits of the LLM's superior performance without compromising data confidentiality.

\section{Operations for Configurable Bricks}
\label{operations}
In the previous section, we introduce the construction of emergent and custom bricks for LLMs. As the number and variety of bricks increases, it becomes crucial to configure LLMs for intricate requirements in real-world applications. This involves utilizing multiple different bricks to execute complex instructions. In this section, we mainly describe several operations associated with the LLMs configurable bricks: routing and retrieving from a vast array of bricks based on instructions ($\S\ \ref{router}$); combining multiple single-purpose bricks to endow the system with composite capabilities ($\S\ \ref{combination}$); updating or refining bricks to align with shifts in world knowledge and requirements ($\S\ \ref{updating}$); growing the bricks to accommodate new capabilities acquired from continuously emerging data ($\S\ \ref{growing}$).

\subsection{Routing and Retrieval}
\label{router}

\begin{figure}
    \centering
    \includegraphics[width=0.95\textwidth]{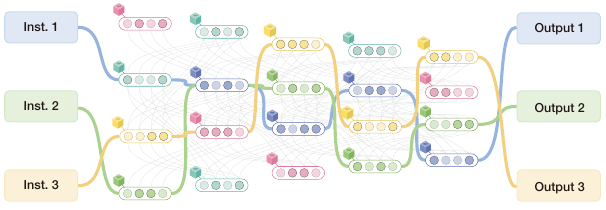}
    \caption{The illustration for brick router and retrieval. It is only necessary to retrieve a subset of bricks to participate in the computation for each instruction.}
    \label{fig:retrieval}
\end{figure}

Given the abundance of emergent and customized bricks, as shown in Figure~\ref{fig:retrieval}, it is essential to establish suitable routing and retrieval methods to utilize these bricks in various situations effectively. 
In this subsection, we first provide an overview of existing routing and retrieval methods, examining them from two perspectives: the categories of bricks and their granularity. 
Then, we engage in a discussion concerning the improvement of current retrieval methods. 

\subsubsection{Emergent Brick Routing}
\label{subsubsec:emergent_brick_route}
Regarding emergent bricks, whether they are defined by humans or self-organized, the main objective of retrieving these bricks is usually to enhance the efficiency of the current model, where only limited parameters are selected for computation. 
Due to emergent bricks being generated during the pre-training, the number of emergent bricks for an LLM is typically limited, often amounting to only a few dozen. Consequently, emergent bricks are selected through a routing function, which assigns a score to each brick based on given instructions or tokens. The bricks with the highest scores are then engaged in the computation process.
By selectively activating the retrieved bricks, it becomes possible to significantly reduce both the training and inference FLOPs, thereby improving computational efficiency. Current routing methods for emergent bricks mainly focus on the pre-defined brick architecture, MoE, and can be categorized into two main categories: trainable route function and fixed route function.

\textbf{Trainable Route Function} In many MoE models, a brick refers to an expert, and a trainable routing function is employed to determine the assignment of each token to its corresponding brick. Typically, the routing function in SwitchTransformer~\citep{DBLP:journals/jmlr/FedusZS22} and GShard~\citep{DBLP:conf/iclr/LepikhinLXCFHKS21} consists of a trainable projection layer, which takes the token representation as input to calculate the gate values for different bricks. The token is then routed to the corresponding expert based on the top-k gate values. However, this method can lead to multiple tokens being assigned to the same brick, resulting in an imbalance of FLOPs among different bricks. To address this issue, alternative approaches have been proposed. One such approach, suggested by \citet{DBLP:conf/icml/LewisBDGZ21}, involves solving a linear assignment problem to route tokens, rather than simply selecting the top-k bricks. \citet{qiu2024layerwise} adopt a recurrent router to establish the associations between routing choices of different layers. \citet{DBLP:conf/nips/ZhouLLDHZDCLL22} propose that bricks select tokens instead of tokens selecting bricks. This method achieves a more balanced distribution of FLOPs among bricks by controlling the number of tokens each brick selects. Additionally, \citet{DBLP:journals/corr/abs-2308-00951} propose the use of soft slots to gather information from all tokens, which are then further processed by different expert bricks. Although these trainable routing functions exhibit meaningful patterns after training~\citep{DBLP:journals/corr/abs-2202-08906}, fully explaining the routing behavior remains a challenging task.

\textbf{Fixed Route Function} Instead of training an unexplainable routing function through the training process, some researchers explore alternative fixed route functions that do not introduce any training parameters. In their work, \citet{DBLP:conf/nips/RollerSSW21} utilize pre-computed hash functions to route tokens to different bricks in a perfectly balanced manner. Similarly, \citet{DBLP:conf/iclr/Zuo00KHZGZ22} find that the behavior of route functions in SwitchTransformer~\citep{DBLP:journals/jmlr/FedusZS22} is akin to random routing. As a result, they suggest employing random routing without any additional parameters. Aside from these random routing methods, \citet{DBLP:conf/naacl/GururanganLHSZ22} propose a token routing approach based on the domains of the current instance, which offers better explainability and encourages specialization among different bricks. However, it should be noted that this method requires domains of nearly equal size to maintain a balance across the different bricks.

\subsubsection{Customized Brick Retrieval}
\label{subsubsec:customized_brick_retrieval}
For customized bricks, the retrieval objective is to enhance LLMs with specific external capabilities that are relevant to the current situation. While retrieval methods are predominantly used in knowledge bricks, as they can be quite extensive, it is crucial to retrieve the most pertinent knowledge from vast sources such as Wikidata.

Several studies~\citep{DBLP:conf/emnlp/FevrySFCK20, DBLP:conf/acl/YeL0S022, DBLP:conf/acl/ZhangZLWYXHLLSZ23} have focused on augmenting LLMs with entity knowledge brick from structured knowledge graphs and have investigated entity linking as a retrieval method to incorporate specific entity brick into the current model. In addition, \cite{DBLP:conf/acl/ChengLC0023} explore encoding Wikipedia into an external memory as knowledge bricks, and they utilize Maximum Inner Product Search (MIPS) to retrieve the most suitable brick for different instances.

In the case of other types of customized bricks, such as task-specific modules, their scale is generally not as vast. Therefore, previous works~\citep{DBLP:conf/emnlp/FriedmanDC21,DBLP:conf/eacl/PfeifferKRCG21, DBLP:journals/corr/abs-2307-13269} have focused on combining and merging all bricks rather than specifically retrieving the most relevant ones. Considering the growing number of task bricks, \citet{DBLP:journals/corr/abs-2402-09997} propose to retrieve and then compose multiple LoRA modules relevant to the input instructions. 

Overall, retrieval methods play a crucial role in incorporating customized bricks, especially knowledge bricks, into LLMs, and further advancements in retrieval techniques can significantly contribute to the effective utilization of these customized bricks.

\subsubsection{Routing and Retrieval Granularity}
\label{subsubsec:retrieval_granularity}

Different routing and retrieval methods can be applied at different levels of granularity, including token-level, sentence-level, and task-level. Each level of granularity offers distinct advantages and considerations.

Token-level routing and retrieval provide greater flexibility and enable precise control over the specific information required for a given task and instance. Many MoE architectures~\citep{DBLP:journals/jmlr/FedusZS22, DBLP:conf/iclr/LepikhinLXCFHKS21, DBLP:conf/nips/ZhouLLDHZDCLL22} employ token-level routing to ensure that different experts acquire more generalized and fundamental capabilities. Additionally, token-level retrieval can be utilized within the context of knowledge brick retrieval to enhance token-level knowledge, such as entity bricks~\citep{DBLP:conf/emnlp/FevrySFCK20, DBLP:conf/acl/YeL0S022, DBLP:conf/acl/ZhangZLWYXHLLSZ23}. However, it is important to note that token-level retrieval and routing can be time-consuming, requiring significant computation and communication costs when applied to every token and every layer.

Sentence-level routing and retrieval provide a higher-level view of the information within a sentence. \citet{DBLP:conf/naacl/GururanganLHSZ22} utilize sentence-level information about the domain of the current instance to route each token to its corresponding domain bricks and promote expertise specialization in specific domains. 
\citet{DBLP:journals/corr/abs-2402-09997} adopt a retrieve-then-compose framework to utilize massive LoRA task bricks, where the model first retrieves several related task bricks based on the input instructions and averages these bricks for final computation.
Furthermore, sentence-level information can also be valuable for text-based knowledge bricks, as the semantic coherence makes the related knowledge of tokens within a sentence highly likely to be the same. \citet{DBLP:conf/acl/ChengLC0023} attempt to construct a sentence-level representation using an average representation of all tokens, which serves as the query to retrieve knowledge brick for the whole sequence understanding.

Task-level routing and retrieval are specifically designed for downstream tasks. Unlike token-level and sentence-level retrieval, which are used during training or inference, task-level retrieval occurs before training and inference. In this approach, retrieval methods aim to identify the most relevant task bricks given all the data associated with a specific task. The retrieved task bricks can then be utilized to perform the task without the need for additional tuning or can serve as a better starting point for subsequent training~\citep{DBLP:conf/eacl/PfeifferKRCG21,DBLP:journals/corr/abs-2307-13269}. 

\subsubsection{Discussion}
\label{subsubsec:retrieval_discussion}
Based on the development of configurable bricks, more advanced routing and retrieval methods need to be proposed.

\textbf{Efficient Routing and Retrieval}
Bricks routing and retrieval during training and inference can be time-consuming due to the increased calculation and communication. \citet{DBLP:journals/corr/abs-2103-13262} show that the training speed of an MoE model can be 3 times slower than a compute-match dense model due to the additional computation and communication on the brick route. Even for customized bricks, some compromise solutions, like retrieval only on specific tokens~\citep{DBLP:conf/emnlp/FevrySFCK20, DBLP:conf/acl/YeL0S022, DBLP:conf/acl/ZhangZLWYXHLLSZ23} or use higher level representation~\citep{DBLP:conf/acl/ChengLC0023}, are proposed to reduce the frequency of routing and retrieval. Better retrieval methods need to be designed to balance accuracy and efficiency.

\textbf{Multi-Level Routing and Retrieval}
Most current retrieval methods are based on single-level retrieval, which may not capture all the information required for accurate retrieval. Additionally, multi-level retrieval is crucial for the collaboration between emergent bricks and customized bricks, as token-level information is empirically used in emergent bricks, while sentence-level and task-level information is more relevant to customized bricks. \citet{DBLP:journals/corr/abs-2212-08066} have made preliminary attempts to incorporate task-level information and token-level information into routing various emergent bricks. However, there is still much to explore in combining multi-level information for retrieval purposes.

\textbf{Active Routing and Retrieval}
Current routing and retrieval methods typically decide in advance where to conduct routing and retrieval based on fixed rules, such as the position of entity mentions or just every token and sentence. We refer to these methods as passive routing and retrieval methods. On the contrary, more proactive methods can allow the LLM to decide where to conduct routing and retrieval during the generation process, which we call active routing and retrieval methods. Active methods can also address the efficiency problem by significantly reducing the frequency of retrieval. \citet{DBLP:conf/emnlp/Zhang00022} have attempted to dynamically augment entity memory when the model generates a specific special token. Furthermore, it is important to investigate whether current LLMs can determine when to retrieve bricks and how to enhance this ability for more advanced configurable foundation models.

\subsection{Combination}
\label{combination}

Single-function bricks often fall short in fulfilling complex instruction demands. Brick combination aims to fuse multiple bricks to possess combined abilities. For example, it enables cross-lingual transfer for the named entity recognition (NER) task to combine a brick trained on English NER datasets with a brick proficient in Chinese capabilities.
In this way, the brick combination can obviate the need to build high-quality annotated datasets for a specific requirement and train models from scratch, thus significantly reducing both human effort and computational costs.

In this section, we divide brick combination methods into two categories based on the operations: parameter weighted average and brick stitching. 
Besides, we also provide a discussion about the future directions for brick combination.

\subsubsection{Parameter Weighted Averaging}
Parameter weighted averaging obtains a merged brick by directly performing an element-wise weighted average of \textit{multiple bricks with the same structure} (homogenous bricks combination). In early research, parameter averaging is applied to the ensemble of models trained from the same task, aiming to boost the model's robustness and performance.
Due to the inherent randomness in the training of deep neural networks and the non-convex nature of loss functions, linearly weighting parameters from two distinct training processes usually fail to yield satisfactory results. 
As a result, many researchers explore the ``mode connectivity'' of deep neural networks~\citep{DBLP:conf/nips/GaripovIPVW18,DBLP:conf/icml/DraxlerVSH18}. These explorations seek to uncover the interconnected paths between the parameters of two models, ensuring that the parameters along this path achieve commendable accuracy.
Further, \cite{DBLP:conf/icml/FrankleD0C20} discover that if two models are initialized from the same well-trained parameters, a straightforward linear weighting can enable the merged model to exhibit superior performance, which inspires subsequent research on parameter average based on pre-trained models. Models fine-tuned from the same pre-trained model, though with different training configurations, including various hyperparameters and data sampling, can be linearly weighted together~\citep{DBLP:conf/icml/WortsmanIGRLMNF22,DBLP:conf/emnlp/QinQYCLHLSZ22,DBLP:conf/eacl/ChronopoulouPFD23,viswanathan2023prompt2model,DBLP:conf/iclr/RuanSMAIFD23}. The resulting merged birck achieves performance comparable or superior to a multi-brick ensemble. Besides, averaging bricks from the same task but different domains has been demonstrated to effectively enhance the domain generalization capacity~\citep{DBLP:conf/nips/ChaCLCPLP21,DBLP:conf/nips/ArpitWZX22,DBLP:conf/nips/RameKRRGC22,DBLP:journals/corr/abs-2208-03306}. Thus, parameter averaging is also employed in efficient federated learning, allowing for a generalized merged brick while simultaneously protecting private data~\citep{DBLP:conf/aistats/McMahanMRHA17,DBLP:conf/iclr/LiHYWZ20,DBLP:conf/icml/KarimireddyKMRS20}.

\begin{figure*}[t]
    \centering
    \includegraphics[width=\textwidth]{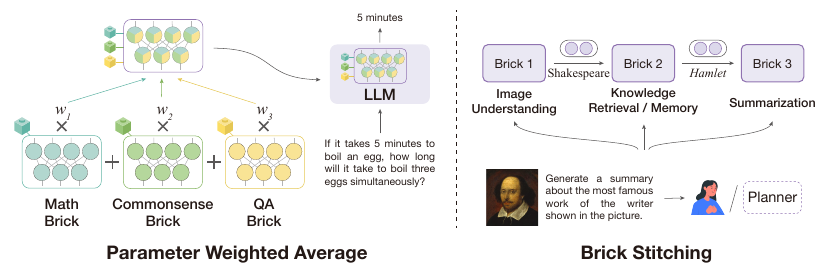}
    \caption{Two widely-used operations for brick combination. (a) Parameter weighted average performs an element-wise average of multiple bricks with the same structures. (b) Brick stitching sequentially concatenates bricks together for complex reasoning.}
    \label{fig:combination}
\end{figure*}

Furthermore, parameter weighted averaging has been introduced to combine bricks from different tasks to facilitate knowledge transfer. In such settings, the contribution of the source task bricks to the target task often varies, which implies that careful design is required to determine the weighting coefficients for the different bricks. \cite{DBLP:conf/nips/MatenaR22} employ Fisher-weighted averaging to transfer capabilities from intermediate tasks to target tasks. \cite{DBLP:journals/corr/abs-2307-13269} leverage combinatorial optimization algorithm to optimize weighting coefficients, aiming to reduce the number of training instances required for the target task. \cite{DBLP:conf/iclr/Jin0P023} determine the coefficients by minimizing the L2 distance between the parameters of merged bricks and source bricks. 
Beyond parameter addition, some researchers discern that subtracting parameters can enable a model to unlearn an undesired capability~\citep{DBLP:conf/iclr/IlharcoRWSHF23}. For instance, \cite{DBLP:journals/corr/abs-2306-14870} perform detoxification by subtracting a brick trained from toxicity instructions. Similarly, \cite{DBLP:journals/corr/abs-2303-17574} mitigate hallucinations by subtracting a brick trained on hallucinated examples.


\subsubsection{Brick Stitching}
Brick stitching involves concatenating several bricks together in sequence based on functional requirements, such that the output of one brick serves as the input for the next. Birck stitching can be applied to combine \textit{bricks with different structures} (heterogeneous bricks combination). Given the substantial discrepancies between features processed by different bricks, a crucial aspect of brick stitching is the training of interaction interfaces between them. This ensures that the outputs from preceding bricks can be effectively interpreted by subsequent ones. While the types of interfaces have been discussed in prior sections (Modality Bricks in $\S$ \ref{plugin}), in this section, our primary focus will be on the way to determine the order and structure of stitched bricks.

\textbf{Heuristic Stitching} 
The manual definition of stitching order based on inference sequence is the most common method of brick stitching. This approach often involves explicitly decomposing a task into several inference steps. For instance, in a visual question-answering task, there's a need to first comprehend the objects and scenes within the given image and then answer questions based on the image content. Correspondingly, concatenating an image encoding model before a language model becomes a widely adopted structure for multi-modal understanding~\citep{DBLP:conf/nips/AlayracDLMBHLMM22,DBLP:conf/icml/0008LSH23,DBLP:journals/corr/abs-2306-13549}. Besides, recent popular LLM-based multi-agent collaboration systems are also based on heuristic brick stitching, where each brick is a whole LLM-based agent and required to solve a subtask, such as front-end design in game development~\citep{DBLP:journals/corr/abs-2307-07924,DBLP:journals/corr/abs-2308-11432,DBLP:journals/corr/abs-2304-12998}.
Heuristic stitching is generally adopted for concatenating model-level bricks, which can independently accomplish specific inference steps. 
In contrast, fine-grained bricks tend to have abstract functions and are usually dependent on surrounding bricks. Arbitrarily concatenating any two bricks typically results in suboptimal collaboration. Recent studies observed that the hidden spaces of two pre-trained models from the same structures and tasks but differing sizes can be linearly transferred~\citep{DBLP:conf/nips/BansalNB21,DBLP:conf/nips/CsiszarikKMPV21}. Based on this insight, \cite{DBLP:conf/cvpr/Pan0Z23} propose an approach that concatenates layer-level bricks from a family of pre-trained models with different sizes. This allows for optimal utilization of computational resources, ensuring maximum performance under given computational constraints.
To further improve the flexibility of stitching different layers, \citet{akiba2024evolutionary} adopt evolutionary optimization algorithms to select optimal composition architectures from predefined stitching search spaces.

\textbf{Planner-based Stitching} 
While heuristic stitching is suitable for tasks with fixed inference steps, real-world instructions often demand varied inference sequences. This implies that we usually need to determine the execution order of different bricks based on the specific given instruction. 
Inspired by early neural module network~\citep{DBLP:conf/cvpr/AndreasRDK16,DBLP:conf/iccv/HuARDS17,DBLP:journals/ijon/Fashandi23}, many brick stitching models are composed of three components: a task planner, which is responsible for decomposing an instruction into several sub-tasks; a controller, which is tasked with generating and receiving the signals for each sub-task and ultimately produces the final answer; multiple bricks, where each brick handles a specific type of sub-task. In such a framework, the stitching order is determined by the task planner.

An intuitive approach is to have the planner generate the execution sequences based on the functional descriptions and usage demonstrations of each brick before performing instruction reasoning~\citep{DBLP:journals/corr/abs-2308-00675,DBLP:journals/corr/abs-2303-17580}. Furthermore, many scholars propose to dynamically decide which brick to call after each reasoning step, allowing the planner to leverage intermediate reasoning results for a more precise determination of the execution sequence~\citep{DBLP:conf/iclr/YaoZYDSN023,DBLP:journals/corr/abs-2306-08640}.
However, a linear execution sequence can be problematic: any error in the selection of bricks can directly impact the final prediction. To address this issue, recent research attempts to incorporate searching strategies for inference, which require the planner to provide multiple options at each step and sequentially test them until a satisfying response is generated~\citep{DBLP:journals/corr/abs-2308-12519,DBLP:journals/corr/abs-2307-16789}.


\subsubsection{Discussion}
Brick combination aims to fuse multiple single-function bricks to fulfill complex instructions. In this section, we delve into parameter-weighted averaging applied to homogenous brick combinations and brick stitching suitable for heterogeneous brick combinations. The essence of parameter-weighted averaging lies in determining the weights for each brick, whereas the key to brick stitching is the alignment of feature spaces across different bricks. While current efforts have initiated preliminary exploration into the brick combination, several challenges remain to be addressed.

\textbf{Combination of Fine-grained Heterogeneous Bricks} 
Most efforts in combination with heterogeneous bricks focus on the integration at the model level. However, parameter redundancies also exist between different models, for instance, both language and image models internally possess neural bricks responsible for understanding real-world concepts~\citep{DBLP:journals/pnas/BauZSLZ020,DBLP:conf/emnlp/GevaCWG22}. Combining fine-grained heterogeneous bricks holds the potential to further reduce parameter redundancies and enhance the reusability of merged bricks across diverse scenarios.

\textbf{A Universal Brick Interaction Interface} 
Within the context of brick stitching, there are two primary interaction interfaces between different bricks: one utilizes discrete, human-readable signals, and the other engages through continuous hidden vectors. The former offers a low training cost but can suffer from information loss; the latter typically results in superior data representation but demands extensive training data and lacks generalizability across different scenarios. Hence, devising a more universally applicable and efficient module interaction method is crucial to enhancing the practicality of brick stitching algorithms.
Besides, a universal interaction interface could boost the scalability of the multi-brick system, allowing bricks capable of the interface to seamlessly stitch with others.

\subsection{Updating}
\label{updating}

The continuous growth and evolution of world knowledge over time presents a unique challenge for LLMs. These models, once trained, may contain outdated information, leading to the phenomenon of hallucination. Therefore, LLMS needs to adapt to shifts in world knowledge. As LLMs become gradually larger, the cost of retraining the model for every new knowledge update request is prohibitively expensive. To this end, methods for quickly updating the knowledge encoded in LLMs have been developed in recent years.
One significant advantage of configurable LLMs is that they allow updating bricks in an isolated manner, which is more efficient (in terms of computation or data) than full parameter fine-tuning. Keeping other parameters frozen may also help minimize unwanted detrimental impacts on other capabilities of the models~\citep{DBLP:journals/corr/abs-2310-16218}. Existing works related to updating bricks have largely focused on editing the knowledge encoded in neural models~\citep{DBLP:journals/corr/abs-2305-13172}. Therefore, the discussion in this section primarily revolves around knowledge bricks updating. However, many concepts for updating knowledge bricks can also be applied to updating other kinds of bricks as well. 
From the perspective of configurable bricks, we can categorize knowledge editing methods into 
(1)~methods that locate and update naturally emergent knowledge bricks, and (2)~methods that inject new customized bricks.

\subsubsection{Locating and Updating Emergent Bricks}

Because of the above challenges, some recent works have explored methods to edit knowledge encoded in the emergent bricks of LLMs. 
To this end, we need to first locate the emergent bricks storing the target knowledge and then update the parameters for knowledge editing.

\textbf{Locating Knowledge Bricks}
Inspired by the hypothesis that FFNs can be regarded as key-value memories~\citep{DBLP:conf/emnlp/GevaSBL21}, existing researches mainly focus on exploring emergent knowledge bricks at the neuron level in the FFN layers. \citet{DBLP:conf/acl/DaiDHSCW22} use integrated gradients~\citep{DBLP:conf/icml/SundararajanTY17} to discover that some factual associations are positively correlated with the activation of a ``knowledge neuron'' in FFNs. This enables the deletion of knowledge by zeroing the activation of knowledge neurons or amplifying knowledge by scaling up the activation. 
\citet{DBLP:conf/nips/MengBAB22} strengthens this hypothesis by causal intervention on activation values and finds that middle-layer FFNs are most responsible for fact recalling. 
\citet{DBLP:journals/corr/abs-2301-04213} show that editing the layers identified by causal intervention does not result in better editing performance. They find that this is because causal intervention identifies which brick carries the target knowledge, but manipulating the parameters of these bricks does not lead to superior performance. It indicates that the knowledge storage of even one entity or tuple involves multiple neurons and defining knowledge bricks at the neuron-group level and layer level may be more suitable.
These works shed light on the challenges of knowledge bricks localization methods.

\textbf{Updating Knowledge Bricks}
After locating the target knowledge neurons, we need to edit the parameters for injecting the updated knowledge. Specifically, the whole FFN layer is regarded as a key-value memory network, and a knowledge neuron is treated as a key-value pair~\citep{DBLP:conf/emnlp/GevaSBL21}. The editing operation is usually performed by replacing the value vector with the target knowledge-enriched representation.
Notably, one emergent knowledge brick usually is responsible for more than one target knowledge, which means knowledge editing usually results in undesirable changes for unrelated knowledge.
To alleviate this issue, extra efforts are made to minimize the impact on unrelated knowledge and capabilities. \citet{DBLP:journals/corr/abs-2012-00363} apply L2 normalization loss on parameter updates to reduce the drift in parameter space. \citet{DBLP:conf/nips/MengBAB22} utilize the layer statistics on a large corpus to directly infer an update that induces a given key-value mapping while minimizing the norm of the weight updates. \citet{DBLP:conf/iclr/MengSABB23} improves upon this by applying multiple key-value mapping at the same time. However, a small change in the parameter space might not translate to a small change in the function space. Hence, many works use a KL divergence loss with adversarial examples to minimize the change of predictions on unrelated inputs after the update~\citep{DBLP:conf/acl/OnoeZPDC23,DBLP:journals/corr/abs-2306-09306,DBLP:conf/iclr/MitchellLBFM22}. Some works also investigate the possibility of using a hyper-network to produce a better parameter update a given gradient information on the representations of a piece of knowledge~\citep{DBLP:conf/emnlp/CaoAT21,DBLP:conf/iclr/MitchellLBFM22}.

\subsubsection{Injecting New Customized Bricks}

An alternative to the locate-and-update paradigm is to inject new bricks that override the existing knowledge. This generally does not require knowledge attribution of neurons or the knowledge update requests are known in advance, since the bricks are created on demand. The main challenges lie in the efficiency of the injection process and its effectiveness in replacing existing knowledge.

Many knowledge-injection methods use full-parameter updates~\citep{DBLP:conf/nips/LewisPPPKGKLYR020,DBLP:conf/iclr/JongZFSC22,DBLP:conf/acl/ZhangHLJSL19,DBLP:conf/coling/SunSQGHHZ20,DBLP:conf/naacl/AgarwalGSA21} and they are typically regarded too compute-heavy for updating knowledge.
One alternative is plug-and-play methods that inject knowledge into a frozen LLM by adding new bricks. \citet{DBLP:journals/corr/abs-2210-04726} propose to train entity embeddings that can be prepended to the hidden states after the input layer. These can be trained on demand, but the update can only happen at the entity level. \citet{DBLP:conf/iclr/HuangSZZR023} build on the work of knowledge neurons and propose to insert and train a new FFN neuron for every knowledge update. \citet{DBLP:conf/icml/MitchellLBMF22} re-route examples concerning updated knowledge to a small model conditioned on a memory of updated knowledge. 
Additionally, a line of works has explored knowledge bricks in the representations instead of the parameters. \citet{hernandez2023inspecting} add an external brick that produces an updated hidden representation that induces the target knowledge when the subject representation is replaced with it. \citet{DBLP:journals/corr/abs-2308-10248} and \citet{DBLP:journals/corr/abs-2310-01405} investigate the possibility of using the difference between the hidden representations of two prompts as steering vectors to induce the target knowledge.

\subsubsection{Discussion}

Currently, updating methods have been largely limited to knowledge bricks. This is because the capabilities of other widely used bricks (see $\S$~\ref{subsubsec:typical-customized-bricks}) such as task and modality rarely require frequent updates. Moreover, these bricks are often trained in isolation without updating the base model. Thus, directly retraining the bricks for the capabilities of interest is typically effective enough. 
Of course, it is not hard to imagine that as we scale up LLMs, even efficient adaptation methods may be too expensive for many applications. Therefore, we believe that exploring more efficient methods for updating tasks, modalities, and other kinds of bricks is a promising future research direction.
Besides, during the deployment process, we often encounter some undesired behaviors of LLMs, such as generating offensive responses when subjected to jailbreaking attacks. Therefore, quickly locating the bricks that lead to the undesired behaviors and correcting them is a promising research direction for efficient alignment.

\subsection{Growing}
\label{growing}

In light of continually growing new knowledge and tasks, the demand for LLM growth thus becomes imperative, necessitating enhancements in the knowledge and capabilities of LLMs while avoiding catastrophic forgetting of existing capabilities. A straightforward strategy to address this challenge involves repetitive training from scratch on both old and new data, which demands prohibitively high costs.
To this end, many efforts have been devoted to continual learning strategies aiming at enabling LLMs to acquire information from new data effectively and efficiently.
In this section, we mainly focus on the strategies for continual learning by increasing the number of bricks. 

\subsubsection{Growing for Pre-Training}
\label{sec:scale-increasing-growth}
Continually learning new knowledge and capabilities from the new pre-training corpus is also important. Moreover, the relationship between the performance of LLMs and model scale has been well-established through scaling laws~\citep{DBLP:journals/corr/abs-2001-08361}. Based on associated observations, increasing the number of emergent bricks presents an effective approach to improving the model performance.

Early works attempt to achieve model continual pre-training by expanding the existing dense parameters, especially, expanding the width, i.e., hidden dimension (neuron-level bricks), and the depth, i.e., the number of layers (layer-level bricks). Then the new bricks and original parameters are trained with the new pre-trained corpus.
An straightforward way to expand the number of parameters is to initialize the expanded parameters with the original parameters and conduct continual pre-training with a mixed corpus~\citep{gong2019efficient,gu2021transformer}. Then many efforts are devoted into further avoiding forgetting old knowledge learned in original parameters and improve the training efficiency.
ELLE~\citep{DBLP:conf/acl/QinZLL0SZ22} enlarges the pre-trained model in width and depth and carefully recovers its capabilities on old tasks through a recovering warmup process. Next, the expanded model undergoes training on a mixture of new data and replayed old data to acquire new information. Similarly, LiGO~\citep{DBLP:conf/iclr/WangPHGKFCWK23} employs a linear mapping of parameters from an existing model to initialize a model with increased width and depth. 
\citet{wu2024llama} frozen the orignal parameters and only train the expanded parameters with new training corpus to avoid forgeting.
In this way, the enlarged LLM can inherit the knowledge of the original small models and thus reduce the costs for continual pre-training. 

An other promising direction is to employ the sparse-activated modular architecture, where only related bricks are selected for computation. Therefore, constructing new bricks and preserving the original bricks frozen will not introduce knowledge forgetting and avoid additional computation costs for inference.
Among these works, the sparse MoE architecture is widely used. The growing operation is performed by increasing the number of experts and keeping the parameters of other layers and experts fixed. During inference, each token will only select the most related experts for computation~\citep{DBLP:journals/corr/abs-2208-03306,DBLP:conf/iclr/KomatsuzakiPLRM23,DBLP:journals/corr/abs-2306-04640,wang2024self}. 


\subsubsection{Growing for Post-Training}
\label{sec:scale-preserving-growth}


Continual learning has been studied for decades for multi-task learning, where the model is required to acquire new knowledge for new task instances and preserve the abilities for existing old tasks~\citep{DBLP:journals/corr/KirkpatrickPRVD16,DBLP:conf/nips/LeeKJHZ17,DBLP:conf/eccv/ChaudhryDAT18,DBLP:conf/eccv/LiH16,DBLP:conf/wacv/ZhangZGLTHZK20}.
Among them, a popular line of research attempts to build an episodic memory, which can be regarded as a memory brick, to store a few representative and informative instances of old tasks~\citep{DBLP:conf/aaai/IseleC18,DBLP:conf/nips/RolnickASLW19,DBLP:conf/naacl/WangXYGCW19,DBLP:conf/acl/HanDGLLLSZ20,DBLP:conf/acl/ZhaoXYG22}. In this way, the model can be continually trained only on instances of new tasks and instances saved in the memory brick, which can save the computational costs. 

Nowadays, benefiting from the plug-and-play characteristic of task bricks, we can continually train LLMs for multiple tasks by constructing a new brick for each new task~\citep{DBLP:conf/acl/MahabadiR0H20,DBLP:conf/cvpr/0002ZL0SRSPDP22,DBLP:conf/eccv/0002ZESZLRSPDP22,DBLP:conf/iclr/RazdaibiedinaMH23}. 
For example, \citet{DBLP:conf/emnlp/MadottoLZMCLYCF21} and \citet{DBLP:journals/corr/abs-2309-14763} increase the model capacity with pluggable Adapter and LoRA modules respectively. To differentiate between the forward propagation routes of new and old data, both works adopt a router to select the appropriate plugin. Besides, constructing new plugins can also introduce more world knowledge, domain knowledge, and complex capabilities for LLMs as discussed in $\S~\ref{plugin}$.


\subsubsection{Discussion}

Based on the insights provided by the aforementioned studies, we suggest choosing the means of brick growing with consideration to the following factors:

\textbf{Task Complexity}
The task complexity serves as a pivotal determinant in the selection of an appropriate model growth strategy. For less complex tasks such as acquiring modest amounts of knowledge, recognizing a new entity category, or manipulating an unseen tool, a viable approach involves growing the model at finer granularities (e.g., introducing a plugin). However, scenarios may arise where a large volume of information should be injected into the model or its knowledge capacity and general performance should be substantially boosted, necessitating growth on a larger scale.

\textbf{Computation Budget}
The computation budget is a rigid constraint on the model scale. While model performance generally improves with growth, the training and deployment costs associated with an expanded model must not exceed the computational budget. Strategies such as plugins, and sparse MoE architectures are representative approaches to model growth within acceptable expenses.

\textbf{Application Targets}
Finally, the growth of the model is always linked to the application targets. For instance, lightweight plugins prove highly advantageous in scenarios emphasizing user-oriented customization. Conversely, when the target is to scale up a model to achieve heightened general AI capabilities, the introduction of a more extensive parameter set through a direct increase in width and depth or other sophisticated architectures (e.g., MoE or progressive networks) becomes imperative.


\section{Empirical Analysis}
\label{experiments}

In previous sections, we discuss that LLMs can be decomposed into emergent bricks and custom bricks from a modular perspective.
Similar to the human brain, neurons in LLMs exhibit the characteristics of sparse activation and function differentiation, meaning each neuron is responsible for specific functionality and is activated when an input instruction requires those functionalities.
Previous works have explored the sparsity~\citep{DBLP:conf/acl/ZhangL00S022,DBLP:conf/iclr/LiYBLRRYCYGK23}, functionality specialization on specific classification tasks~\citep{DBLP:conf/emnlp/WangWZ0LL22}, and modular grouping~\citep{DBLP:conf/acl/ZhangZLXW00XS023} on encoder model, BERT~\citep{DBLP:conf/naacl/DevlinCLT19}, or encoder-decoder model, T5~\citep{DBLP:journals/jmlr/RaffelSRLNMZLL20}. In this paper, we focus on the analysis of widely-used decoder-only models with instruction-following chat data.

Specifically, we conduct a detailed analysis of the following questions: (1) \textbf{Are LLMs sparsely activated}, meaning that only a few neurons influence the final output when processing each token? (2) \textbf{Do neurons exhibit functional specialization}, with their activation values highly correlated to the capabilities required by the instruction? (3) \textbf{Do LLMs have the potential to be modularly split}, which means that different capabilities activate different partitions of neurons? In this section, we present our empirical analysis, beginning with an introduction to the formal definition of neurons and their activation values, the functionality localization of neurons, and finally, we present the experimental results.

\subsection{Functionality Localization}
In this section, we attempt to the analysis of neurons in the feedforward layers. Previous works indicate that feedforward layers in Transformer can be regarded as key-value memory networks~\citep{DBLP:conf/emnlp/GevaSBL21} and provide world knowledge for sequence understanding. Therefore, we mainly focus on the feedforward layers for analysis.

\paragraph{Neurons and Activations}
The feedforward layers (FFNs) employ two-layer projections or gated projections for each token in the sequences. The calculation can be written as $\text{FFN}(\mathbf{x}) = \text{FFN}^O(\text{FFN}^I(\mathbf{x})) = \mathbf{W^O}(\text{FFN}^I(x)) + \mathbf{b^O}$. Here, $\mathbf{W^O} \in \mathbb{R}^{d \times d_{ff}}$ and $\mathbf{b^O} \in \mathbb{R}^d$ are the weight matrix and bias vector for the output linear layer $\text{FFN}^O(\cdot)$. As for $\text{FFN}^I(\cdot)$, there are two variants:
\begin{align*}
    \text{Vallina FFN:} \quad &\text{FFN}^I(\mathbf{x}) = \sigma \left(\mathbf{W^I}\mathbf{x} + \mathbf{b^I}\right), \\
    \text{Gated FFN:} \quad &\text{FFN}^I(\mathbf{x}) = \sigma\left(\mathbf{W_G}\mathbf{x} + \mathbf{b_G}\right) \odot \left(\mathbf{W^I}\mathbf{x} + \mathbf{b^I}\right).
\end{align*}
Here, $\mathbf{W_G}, \mathbf{W^I} \in \mathbb{R}^{d_{\text{ff}}\times d}$ and $\mathbf{b^I}, \mathbf{b_G} \in \mathbb{R}^{d_{\text{ff}}}$ are the weight matrices and bias vectors for the input linear layer $\text{FFN}^I(\cdot)$ and gate linear layer $\text{FFN}_G(\cdot)$. Following previous works~\citep{DBLP:conf/acl/ZhangZLXW00XS023,DBLP:conf/emnlp/WangWZ0LL22}, we can split an FFN layer as $d_{\text{ff}}$ neurons, each consisting of a row in the input and gate layer as well as a column in the output layer. The outputs of FFN layers can be rewritten as the sum of all neuron outputs: $\text{FFN}(\mathbf{x}) = \sum_i^{d_{\text{ff}}} \text{FFN}^I(\mathbf{x})_i \mathbf{W}^O_{:,i} + \mathbf{b}^O_i$. We define the intermediate output, $\text{FFN}^I(\mathbf{x})_i$, as the activation value of $i$-th neuron. Intuitively, if the magnitudes of activation values are small, then the corresponding neuron will have a limited impact on the final outputs and vice versa. Therefore, the activation values are widely used as indicators for the functionality of neurons.

\paragraph{Functionality Score}  
In the following paragraphs, we will introduce how to locate neurons with specific functionalities. As mentioned before, the activation values can reflect the contributions of each neuron to the FFN layer output and thus are usually used as the indicator for the functionality. Then we will present the process to calculate the functionality scores of neurons on given functionalities.

\begin{table*}[]
    \centering
    \setlength{\extrarowheight}{3pt}
    \caption{The functionalities and their corresponding data labels used in this paper. The data labels in Infinity-Instruct are not mutually exclusive, meaning that a single instance can belong to multiple different data labels.}
    \begin{tabular}{l|p{11.5cm}}
    \toprule
    Functionality & Data Labels  \\ \midrule
    Coding   & Python Programming, SQL Programming, Java Programming, C++ Programming, Javascript Programming, C\# Programming, Object-oriented Programming, Code Comments, Code Writing \\ \midrule
    Math     & Mathematical Reasoning, Mathematical Modeling, Basic Mathematics, Mathematical Analysis, Mathematical Applications, Mathematical Proof, Mathematical Explanation, Mathematical Concept Explanation, Solving Complex Mathematical Problems, Basic Mathematics Calculations \\ \midrule
    Linguistic & Sentence Structure Analysis, Syntactic Understanding, Linguistic Knowledge, Syntactic Generation, Syntactic Analysis \\ \midrule
    Knowledge & Health Knowledge, Geographic Knowledge, General Knowledge about Science, Legal Knowledge, Physics Knowledge, Chemistry Knowledge, Literary Knowledge, Sociology Knowledge, Popular Science Knowledge, Biology Knowledge, Astronomy Knowledge, Psychological Knowledge, Economic Knowledge, Clinical Medical Knowledge, Environmental Knowledge, Religious Studies Knowledge, Geometry Knowledge \\ \midrule
    Translation & Multilingual Translation, Translation Ability, Chinese English Translation, Machine Translation, French Translation \\ \midrule
    Ethics and Moral & Ethical Judgment, Ethical Reasoning, Ethical Analysis, Ethical and Moral Reasoning, Ethical Thinking, Ethical Guidance, Unethical Behavior Simulation, Unethical Behavior, Ethics and Morality, Moral Standards \\ \midrule
    Writing & Scriptwriting, Creative Writing, Narrative Writing, Technical Writing, Writing Guidance, News Writing, Script Writing, Creativity Writing, Product Description Writing, Screenwriting Ability
    \\ \bottomrule
    \end{tabular}
    \label{tab:data_type}
    \vspace{-1em}
\end{table*}

Specifically, we denote the functionality, such as coding ability, as $f$, a collection of chat instances as $\mathcal{C} = \{(p_0, r_0), ... , (p_n, r_n)\}$, where $p_i$ and $r_i$ are user input prompt and model-generated response. We define the functionality label for each chat instance $(p_i, r_i)$ as $y_i^f$, where $y_i^f = 1$ when $p_i$ requires $\mathcal{M}$ to have the capability $f$ to generate the correct answer; otherwise, $y_i^f = 0 $. For example, given the prompt $p=``\textit{How can we select unique elements from a list in Python?}''$, its functionality label for code ability is $1$ and its functionality label for translation ability is $0$.

We denote the LLM requiring to be analyzed as $\mathcal{M}$, which has $L$ Transformer layers and $L\times d_{ff}$ neurons. Given a neuron $n$, as the FFN layers are computed in a token-wise manner, we can collect the activation values of the neuron $n$ on the collection $\mathcal{C}$. For the instance, $(p_i, r_i)$, there are $l_i$ activation values, and we define the collection of the absolute value of activation values as: $\mathcal{A}_i = \{|a^i_0|, ..., |a^i_{l_i}|\}$ from the $l_i$ tokens in $p_i$. We define the activation value of neuron $n$ on the instance $(p_i, r_i)$ as the average value of $\mathcal{A}_i$. Then, following \cite{DBLP:conf/acl/ZhangZLXW00XS023}, we use the average precision score as the functionality score of $n$ on the functionality $f$:
\begin{equation}
    \text{FuncScore}(n, f) = \text{AvgPrecision}(\{\overline{\mathcal{A}}_0, ..., \overline{\mathcal{A}}_i\}, \{y_0^f, ..., y_n^f\}).
\end{equation}
Intuitively, a higher functionality score suggests a stronger correlation between neuron $n$ and capability $f$. That is to say, if the $\text{FuncScore}(n, f)$ is high, neuron $n$ exhibits higher activation levels when the input prompt necessitates capability $f$ and lower activation when it does not.

\begin{figure*}
    \centering
    \begin{subfigure}[b]{0.242\textwidth}
    \centering
    \includegraphics[width=\textwidth]{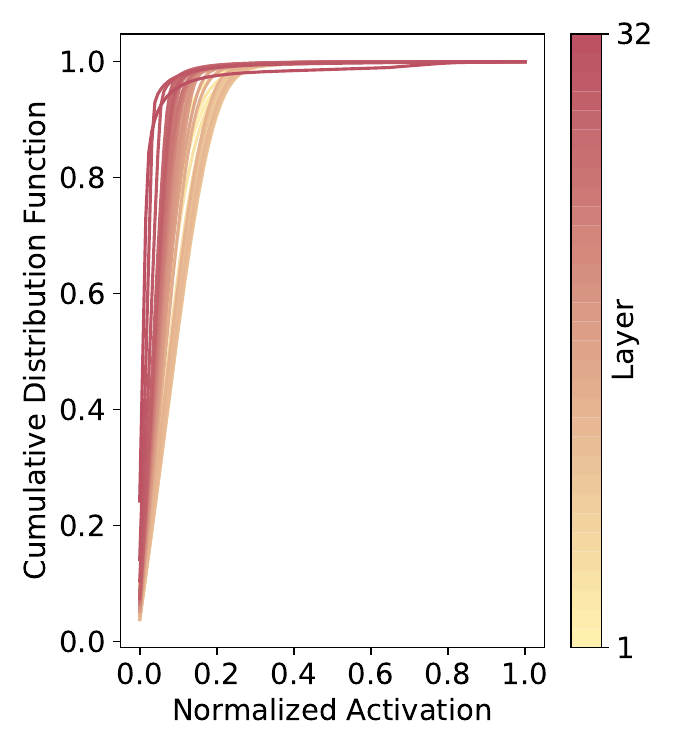} 
    \caption{}
    \end{subfigure}
    \begin{subfigure}[b]{0.237\textwidth}
    \centering
    \includegraphics[width=\textwidth]{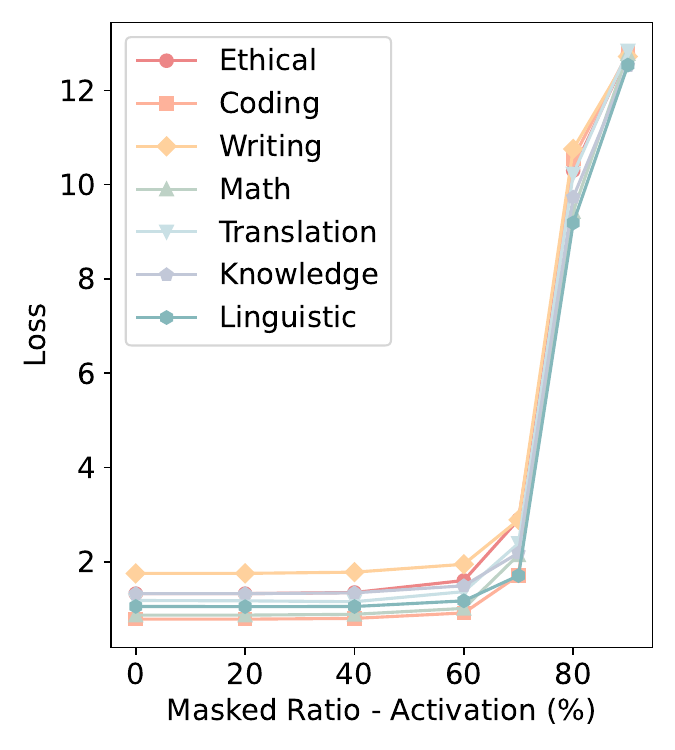} 
    \caption{}
    \end{subfigure}
    \begin{subfigure}[b]{0.242\textwidth}
    \centering
    \includegraphics[width=\textwidth]{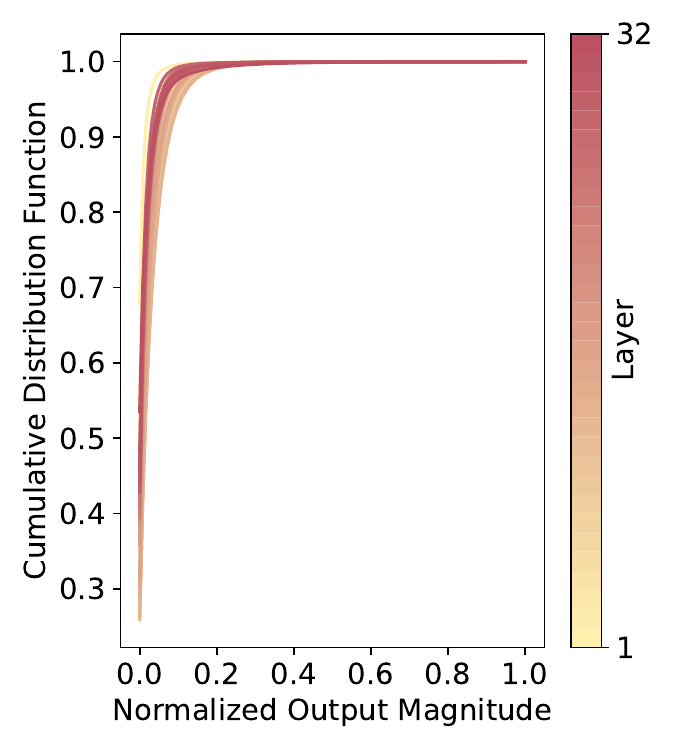} 
    \caption{}
    \end{subfigure}
    \begin{subfigure}[b]{0.237\textwidth}
    \centering
    \includegraphics[width=\textwidth]{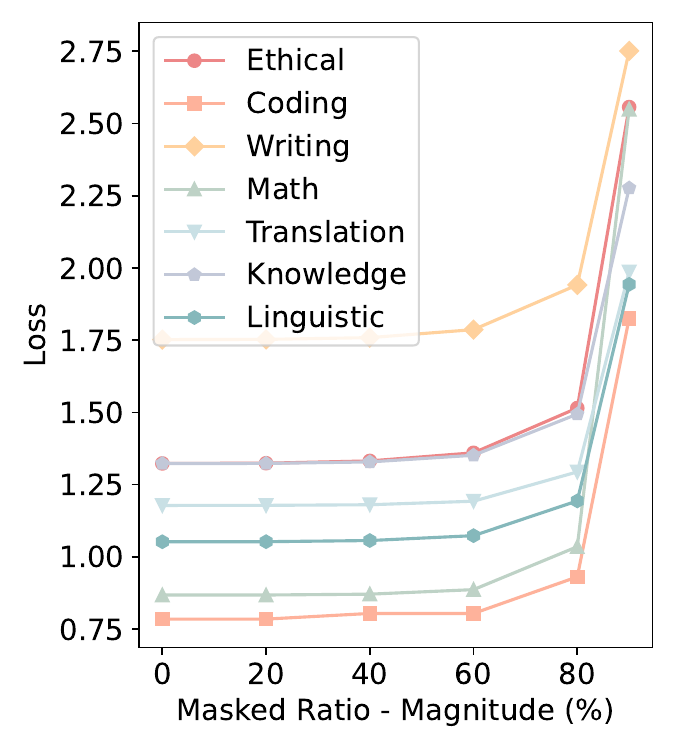} 
    \caption{}
    \end{subfigure}
    \caption{Sparsity activation for neurons in Llama-3-8B-Instruct. (a)(c)~The cumulative distribution function of normalized activation values and output magnitudes. (b)(d) Impact of neurons with low activation values and output magnitudes. }
    \label{fig:sparse}
\end{figure*}

\begin{figure*}
    \centering
    \begin{subfigure}[b]{0.242\textwidth}
    \centering
    \includegraphics[width=\textwidth]{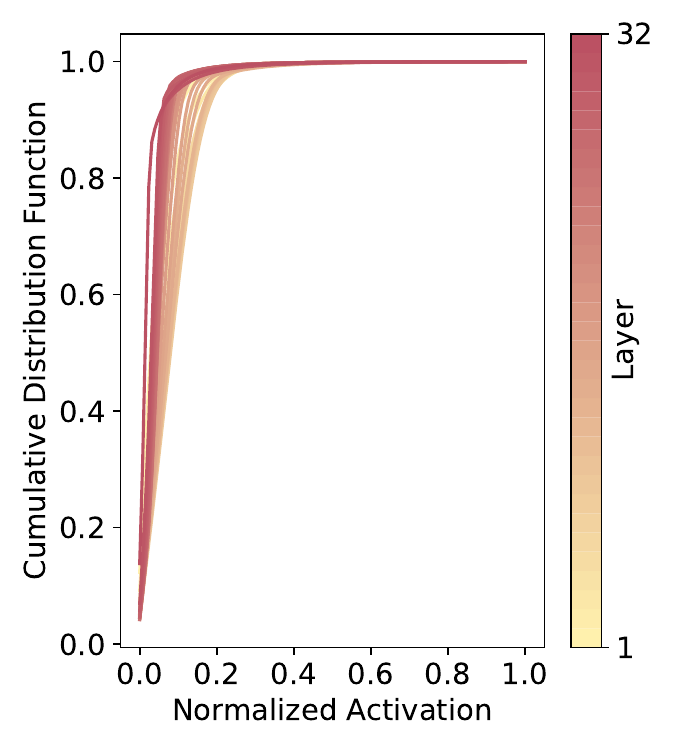} 
    \caption{}
    \end{subfigure}
    \begin{subfigure}[b]{0.237\textwidth}
    \centering
    \includegraphics[width=\textwidth]{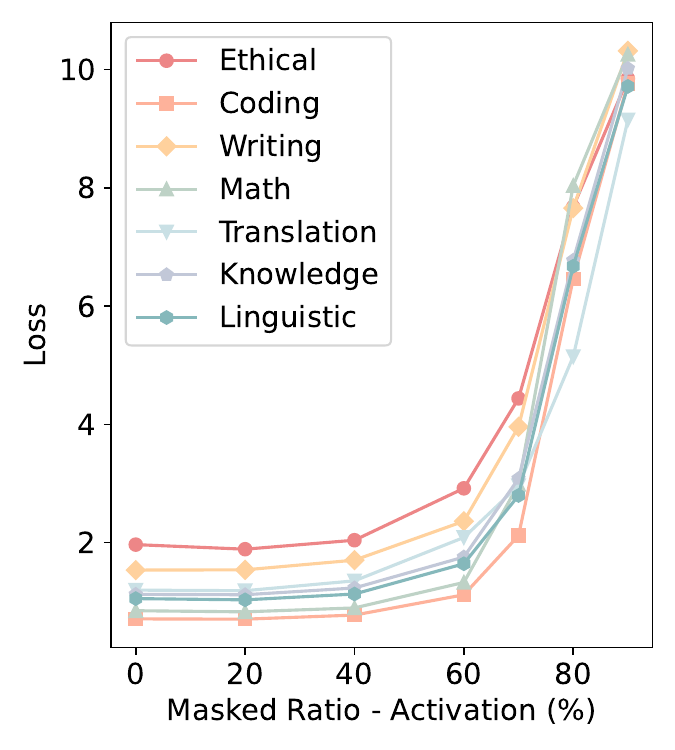} 
    \caption{}
    \end{subfigure}
    \begin{subfigure}[b]{0.242\textwidth}
    \centering
    \includegraphics[width=\textwidth]{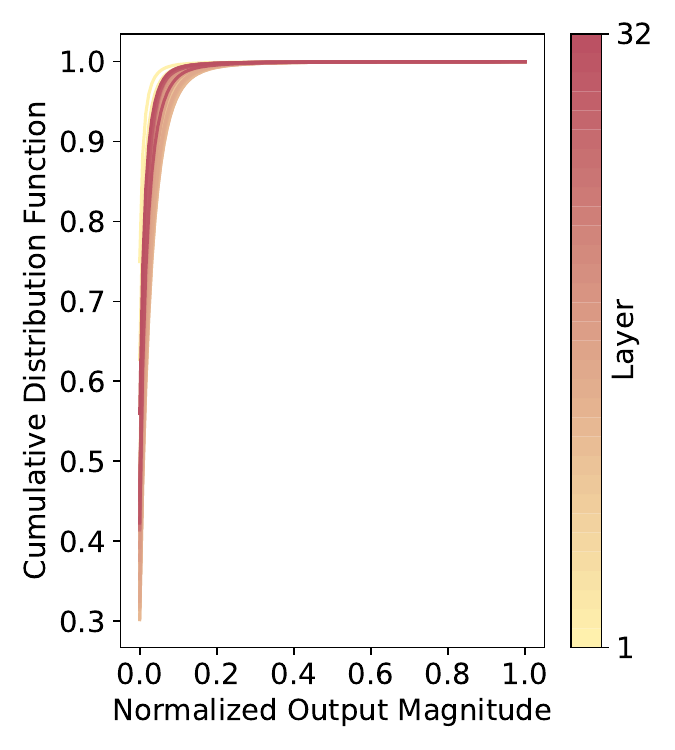} 
    \caption{}
    \end{subfigure}
    \begin{subfigure}[b]{0.237\textwidth}
    \centering
    \includegraphics[width=\textwidth]{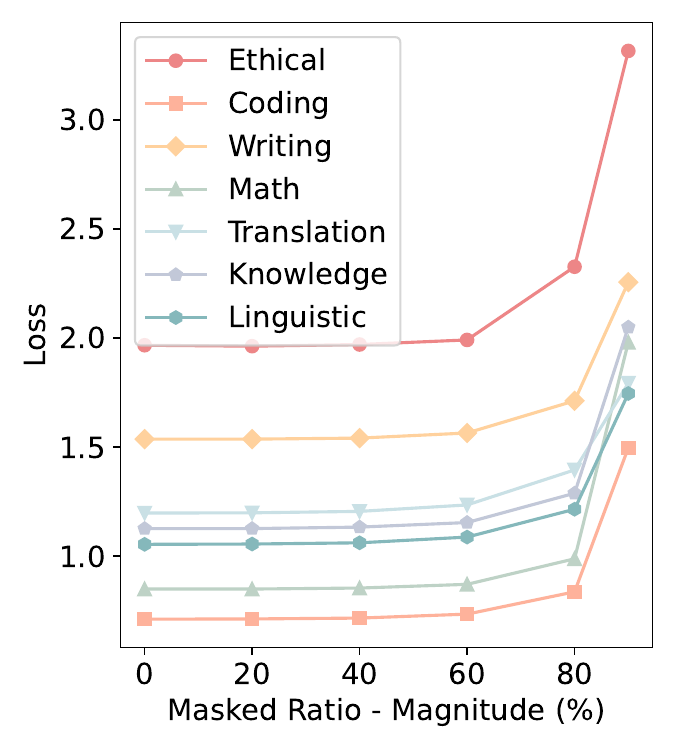} 
    \caption{}
    \end{subfigure}
    \caption{Sparsity activation for neurons in Mistral-7B-Instruct-v0.3. (a)(c)~The cumulative distribution function of normalized activation values and output magnitudes. (b)(d) Impact of neurons with low activation values and output magnitudes. }
    \label{fig:sparse2}
\end{figure*}

\begin{figure*}
    \centering
    \begin{subfigure}[b]{0.98\textwidth}
    \centering
    \includegraphics[width=\textwidth]{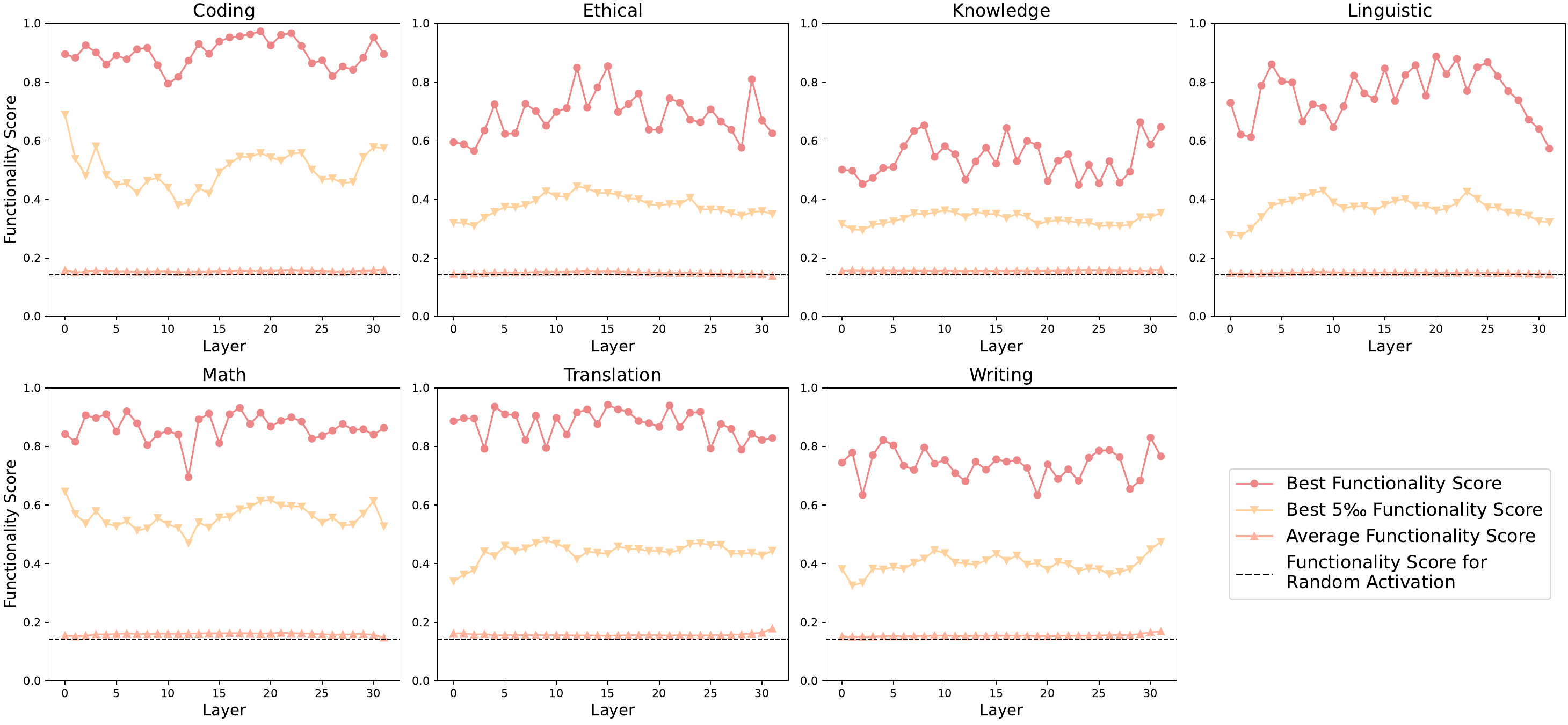} 
    \caption{Llama-3-8B-Instruct}
    \end{subfigure}
    \begin{subfigure}[b]{0.98\textwidth}
    \centering
    \includegraphics[width=\textwidth]{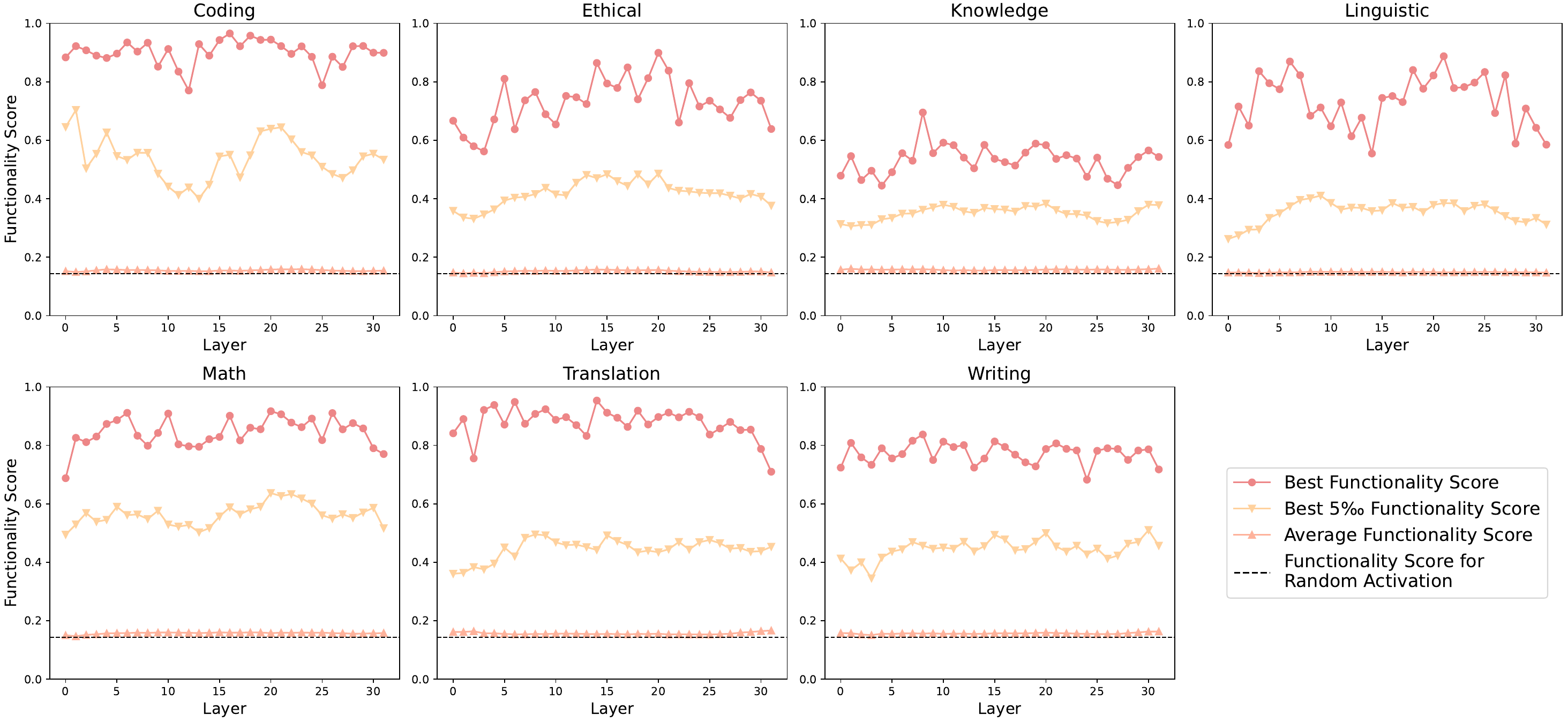} 
    \caption{Mistral-7B-Instruct-v0.3}
    \end{subfigure}
    \caption{Distribution of functionality scores across different layers. We also present the average functionality scores and functionality scores for random activation.}
    \vspace{-1em}
    \label{fig:funcscore}
\end{figure*}
\subsection{Experimental Settings}

To analyze the functionality specialization and partition, we need a dataset annotated with abilities required by each instance. Therefore, in the experiments, we adopt the Infinity-Instruct dataset~\footnote{\url{https://huggingface.co/datasets/BAAI/Infinity-Instruct}}. 
Each instance in Infinity-Instruct consists of user prompts, model-generated responses, and several abilities required by the given user prompts. There are thousands of different abilities across the entire dataset and we summarize $7$ typical and widely-used functionalities and its corresponding data labels for our analysis. The detailed functionalities and data labels are listed in Table~\ref{tab:data_type}. (1)~\textbf{Coding}: Specializes in a variety of programming languages including Python, Java, and C++, with expertise in object-oriented programming and effective code documentation. (2)~\textbf{Math}: Encompasses a wide range of mathematical skills, from basic calculations to complex problem-solving and theoretical proofs. (3)~\textbf{Linguistic}: Focuses on the analysis and generation of syntactic structures, enhancing understanding and applying linguistic knowledge effectively. (4)~\textbf{Knowledge}: Covers an extensive array of subjects such as science, literature, and religion, reflecting the deep understanding and application of specific disciplinary knowledge. (5)~\textbf{Translation}: Showcases multilingual capabilities, specializing in Chinese-English translations among other languages, with proficiency in machine translation systems. (6)~\textbf{Ethics and Moral}: Concentrates on ethical reasoning and judgment, exploring concepts from ethical analysis to the implications of unethical behaviors and moral standards. (7)~\textbf{Writing}: Spans a variety of genres and formats including scriptwriting, creative writing, and technical documentation, emphasizing creativity and clear communication.

We manually select data labels that meet our specific functionality requirements. Since each instance may demand various abilities, our goal is to analyze the specialization and distribution of functionality across different abilities. Therefore, we retain only those instances that belong exclusively to one of the aforementioned types. We randomly sample $1,000$ instances from each data type for further analysis.

As for the backbone model, we adopt two widely used models, Llama-3-8B-Instruct and Mistral-7B-Instruct-v0.3, for analysis. Both two models are trained with large-scale corpus and chat data.

\subsection{Sparse Activation}
In this subsection, we focus on the first question: are LLMs sparsely activated, just similar to human brains? If LLMs are sparsely activated, we can select part of them for the computation of each input to reduce the computational costs. To evaluate the sparsity of LLMs, we attempt to calculate the distribution of neuron activations and the impact of the neurons with low activation values. Besides, as the activation values are intermediate results of FFN, following \cite{zhang2024relu}, we directly observe the impact of each neuron on the output, i.e., output magnitude, for sparsity evaluation. Specifically, for $i$-th neuron in FFN, the output magnitude is defined as $||\text{FFN}(x)_i||_2$. For the convenience of introduction, we term the activation values and output magnitudes as indicators.
Then, to further evaluate the impact of these neurons with low indicators on the model performance, we further assess the loss variation with these neurons masked. Specifically, given an input sequence, we first compute the intermediate indicators for all tokens and in all FFN layers. Then we can select and mask $k$\% neurons with the lowest indicators for each token and each layer.

The results are shown in Figure~\ref{fig:sparse} and Figure~\ref{fig:sparse2}. The results indicate that: 
(1)~The normalized indicators for $80\%$ neurons are lower than $0.2$. It indicates that the impact of neurons is present in a long-tail distribution, and only a few neurons have a significant impact on the output of FFN layers.
(2)~The distributions are similar across different layers and two decoder-only models with different training data. It proves that the sparsity issues widely exist in LLMs.
(3)~Across all data from the seven functionalities, when the proportion of masked neurons is low ($70\%$ when using activation values as indicators, $80\%$ when using output magnitudes as indicators), there is almost no decline in the model performance. This further demonstrates that similar to human brains, LLM exhibits sparsity in its parameter usage. This suggests the potential to significantly reduce computational costs without compromising performance by using part of the parameters for each input.
(4)~Compared to activation values, using output magnitudes as indicators results in a greater number of neurons with normalized indicators below $0.2$, allowing more neurons to be masked without a performance drop. This indicates that output magnitudes can achieve a higher degree of sparsity. This finding encourages further research to identify more effective indicators for assessing neuron usefulness.

\begin{figure*}
    \centering
    \begin{subfigure}[b]{0.48\textwidth}
    \centering
    \includegraphics[width=0.9\textwidth]{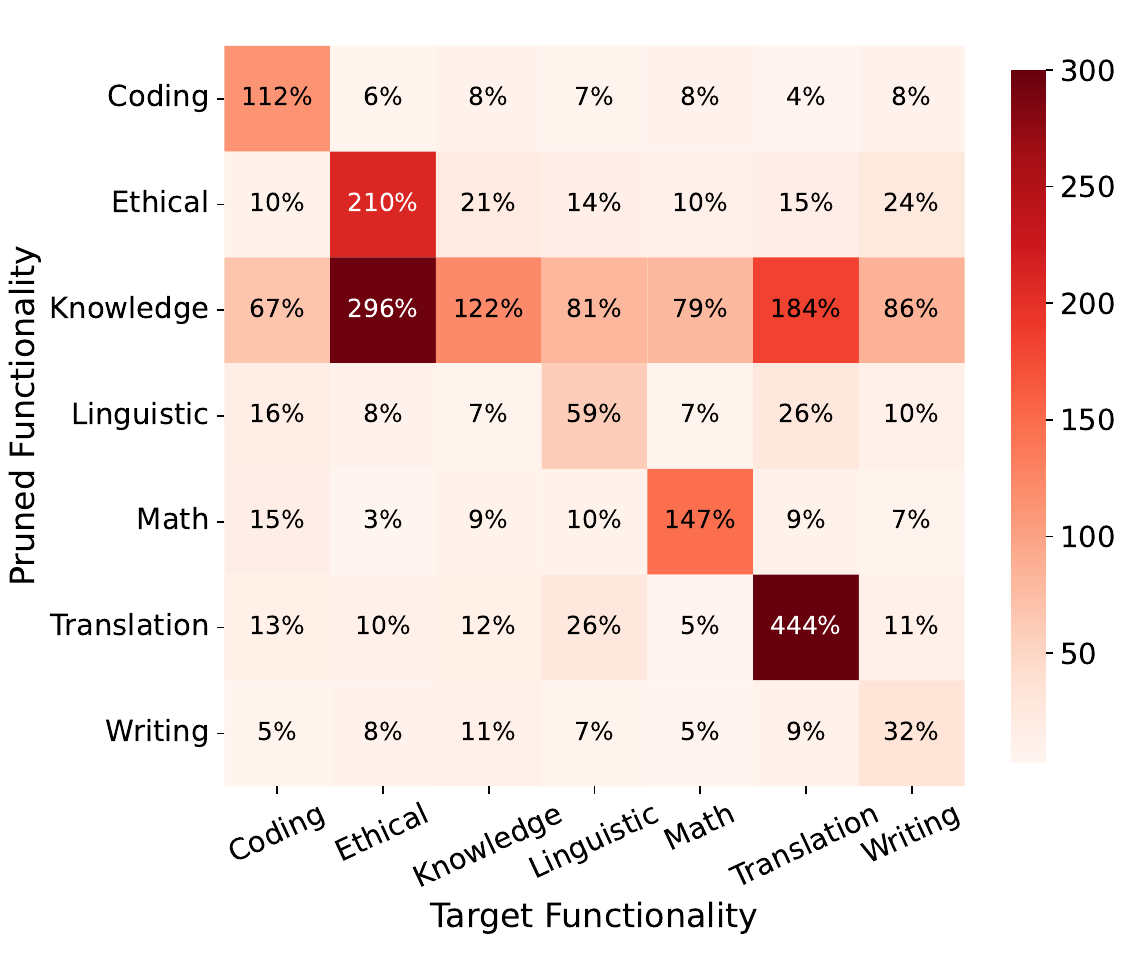} 
    \caption{Llama-3-8B-Instruct}
    \end{subfigure}
    \begin{subfigure}[b]{0.48\textwidth}
    \centering
    \includegraphics[width=0.9\textwidth]{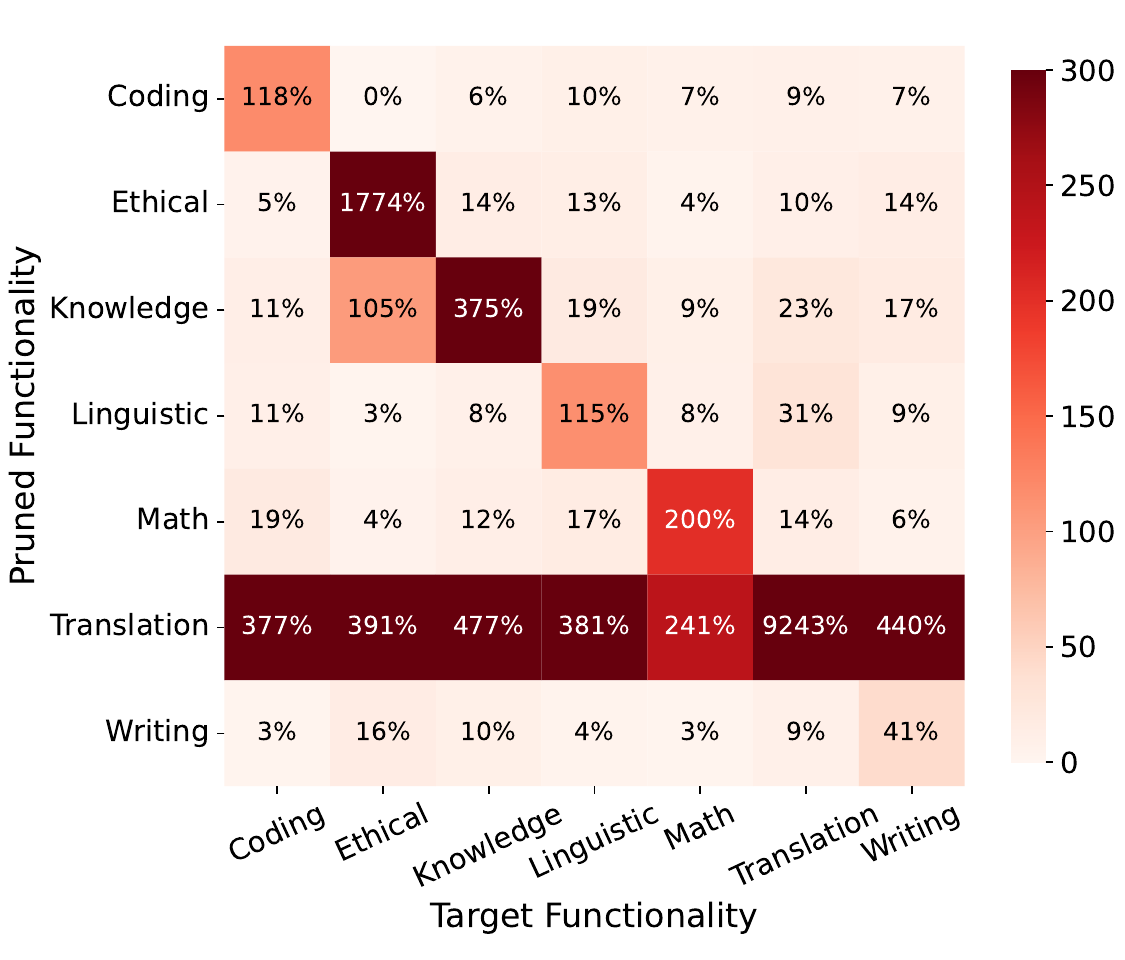} 
    \caption{Mistral-7B-Instruct-v0.3}
    \end{subfigure}
    \caption{Increase in perplexity after pruning neurons for specific functionality: $\frac{\text{PPL}_{\text{pruned}} - \text{PPL}_{\text{origin}}}{\text{PPL}_{\text{origin}}}$ (\%).}
    \vspace{-1em}
    \label{fig:ppl}
\end{figure*}

\subsection{Functionality Specialization}

In this subsection, we discuss the functionality specialization of neurons in LLMs.
First, we attempt to locate neurons corresponding to different functionalities within LLMs. We calculate the functionality scores of all neurons across each layer for seven functionalities. As shown in Figure~\ref{fig:funcscore}, we present the best functionality scores for neurons in each layer. We also provide the functionality scores of randomly activated neurons and the average functionality scores of all neurons in a single FFN layer as baselines. From the results, we can observe that: (1) The best functionality scores across these seven functionalities are significantly higher than those of randomly activated neurons and the mean functionality scores. This demonstrates that each FFN layer contains neurons highly associated with these seven functionalities, indicating that these neurons have differentiated into distinct functionalities during pre-training and alignment processes. (2) The functionalities of Coding, Math, and Translation can achieve functionality scores higher than $0.8$ in most layers. This suggests that neurons associated with these three functionalities are more specialized compared to the other four functionalities. Instructions for these functionalities require LLMs to understand or generate sequences distinctly different from the English natural language, hence the neurons activated show high specificity.
(3)~We also present the $5$\textperthousand{} highest functionality scores across different layers. We can observe that there are large gaps between the best functionality scores and the best $5$\textperthousand{} scores. As mentioned before, the neurons are sparsely activated and there are only a few neurons are highly associated with each specific functionality. Thus, the functionality scores drop quickly with the increase of the number of neurons.
(4)~The average functionality scores are almost equivalent to the functionality scores of randomly activated neurons, indicating that the functionality scores of most neurons are close to the random baseline. This further suggests that for each capability, only a small subset of neurons is highly associated with it.

Besides, inspired by \cite{DBLP:conf/acl/ZhangZLXW00XS023}, we further conduct a perturbation study to verify the functionality specialization of neurons with high functionality scores. In this study, given the specific functionalities, we prune $5\%$ neurons with high functionality scores and evaluate the pruned models on all data from $7$ functionalities. For instance, given the coding functionality, we manually set the activation value of the neurons with high functionality scores as zero and evaluated the impact of the pruned neurons on all functionalities. Theoretically, if the pruned neurons are highly specialized to specific functionality, they are supposed to only have an impact on the corresponding functionality and have minor impacts on other functionalities. As all data used in our experiments are generation tasks without a clear task format, we adopt perplexity as our evaluation metric. The results of the perturbation study are shown in Figure~\ref{fig:ppl}. From the results, we can observe that:
(1)~Values on the diagonal are generally higher than those of the diagonal. This indicates that after pruning neurons for specific functionality, the model’s performance significantly deteriorates in the corresponding functionality while having less impact on other functionalities.
(2)~Pruning neurons for knowledge in Llama and neurons for translation in Mistral significantly affect all other functionalities. This may be due to the presence of resident neurons in the FFN~\citep{song2023powerinfer}, which are frequently activated for most inputs. Including these neurons when selecting for functional specificity results in a substantial impact on model performance. In the future, we will explore more effective methods to locate function-specific neurons.

\begin{figure*}
    \centering
    \begin{subfigure}[b]{0.48\textwidth}
    \centering
    \includegraphics[width=0.9\textwidth]{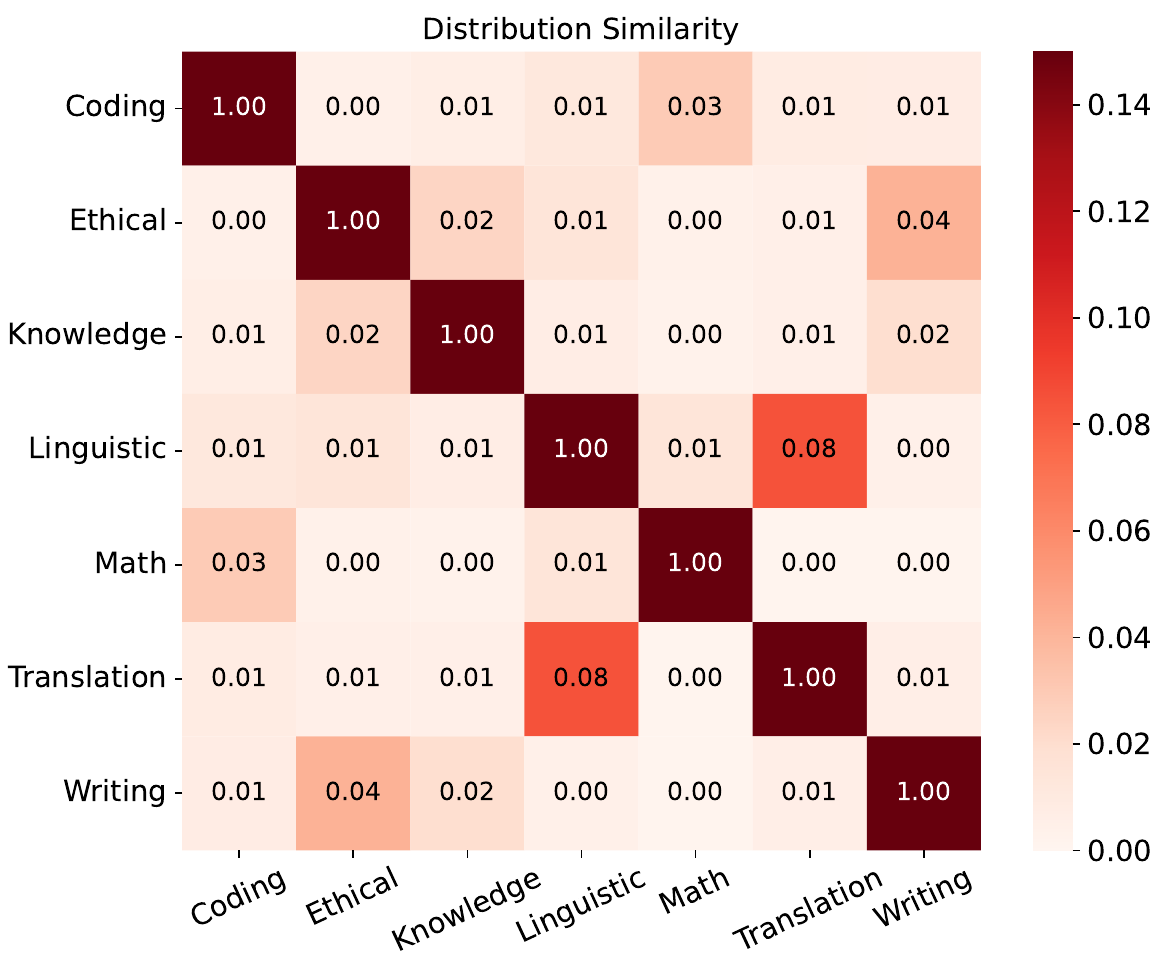} 
    \caption{Llama-3-8B-Instruct}
    \end{subfigure}
    \begin{subfigure}[b]{0.48\textwidth}
    \centering
    \includegraphics[width=0.9\textwidth]{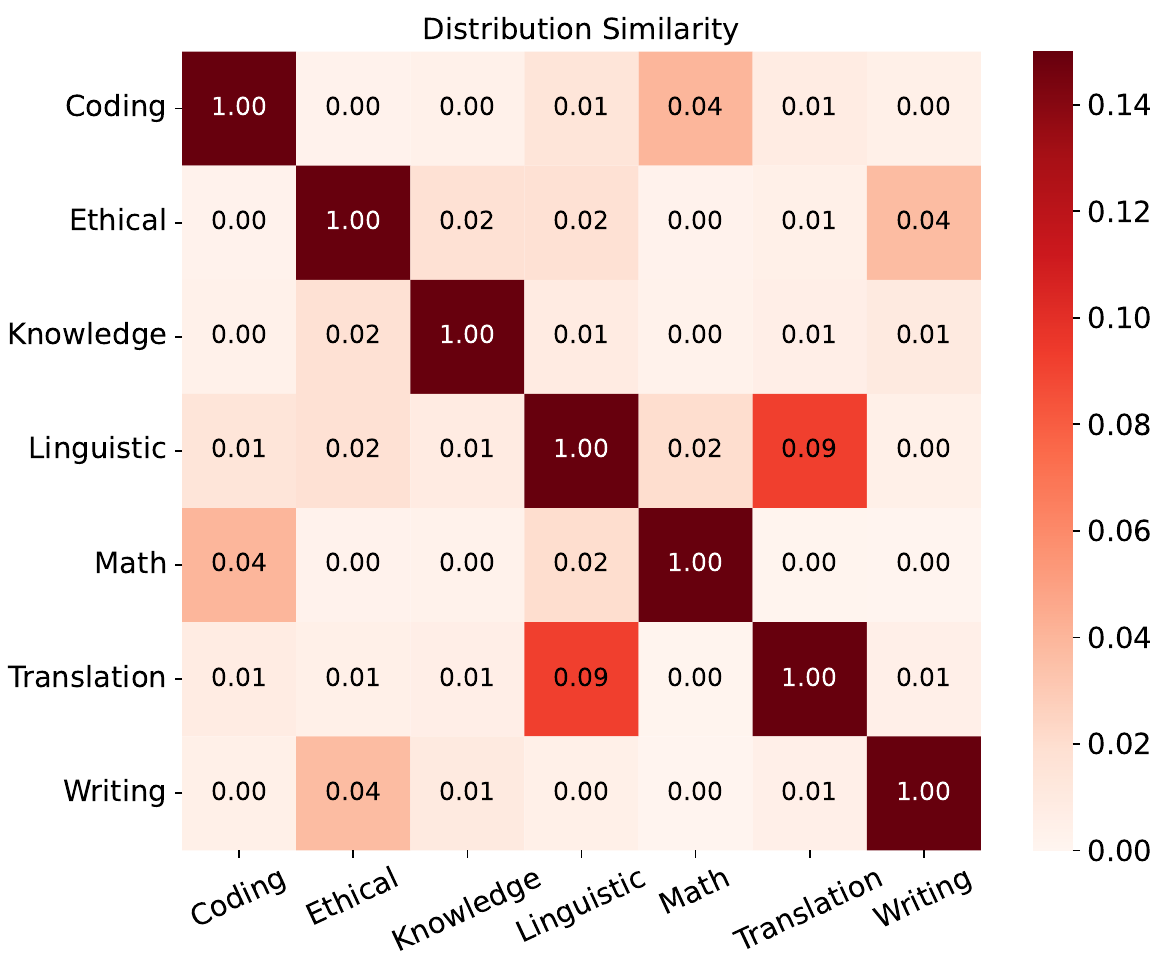} 
    \caption{Mistral-7B-Instruct-v0.3}
    \end{subfigure}
    \caption{Distribution similarity between specialized neurons of different functionalities, where the similarity between two randomly selected neuron groups are $0.05$.}
    \vspace{-1em}
    \label{fig:distribution}
\end{figure*}

\subsection{Functionality Partition}
From experimental results in previous subsections, we can observe that neurons are sparsely activated, and each neuron exhibits specific functionalities. Based on these observations, in this sub-section, we further explore the potential for modular partitioning in LLMs. Similar to the human brain, neurons can be divided into several regions, each region containing neurons specialized for specific capabilities, collaborating yet not interfering with each other. Therefore, in this subsection, we attempt to visualize whether there are distinct partitions within the LLMs across the aforementioned seven capabilities.

Specifically, we compute the distribution similarity between top-$5\%$ neurons of different functionalities. The results are shown in Figure~\ref{fig:distribution}. From the results, we can observe that: (1)~Values on the diagonal significantly outperform those off the diagonal, indicating that neurons for different functionalities are distinctly different. 
(2)~The similarity between neurons for translation and linguistic functionalities is greater than the random value, which is due to the need to ensure grammatical correctness in the output language during the translation process. In the future, an important research direction is to explore how to accurately cluster different neurons into distinct groups. This approach could avoid the need to select parameters at a neuron level.

\section{Open Problems and Future Directions}
\label{discussion}

\subsection{Correlation between Emergent and Customized Bricks}
\label{correlation}
%


The emergent and customized bricks are the essence of configurable foundation models that make the training and updating of LLMs more flexible and scalable. Configuring LLMs with both emergent and customized bricks can promote decomposing and recombining functionalities for existing LLMs. However, as these two types of bricks acquire capabilities through different stages, there exist subtle discrepancies between their properties. For instance, emergent bricks can learn some outdated factual knowledge from the pre-training corpus, while customized bricks post-processed with updated documents may have the latest but also overlapped knowledge. This could lead to unexpected collisions and redundancy in their functionalities, resulting in potential performance degradation and extra computation costs. We advocate further efforts to better manage the integration and cooperation of emergent and customized bricks for ensuring optimal performance and efficiency in configurable LLMs.

\textbf{Construction} The potential collision and redundancy between the functionalities of emergent and customized bricks can be traced back to their construction process. Though emergent bricks can be human-defined or self-organized, their capabilities are attained through the large-scale pre-training procedure, which is typically conducted in an end-to-end manner, making them relatively hard to interpret and localize. For adaptations to new tasks and knowledge that the existing model does not have, customized bricks are constructed after the pre-training stage with delicately designed structures and learning objectives. However, since it is impossible to enumerate the capabilities and knowledge of existing models, incorporating multiple customized bricks for new capabilities and knowledge can also introduce redundancy and collision.

In addition, the granularities of both emergent and customized bricks have several variations and each of them may possess distinct abilities at different levels. The diverse combinations of emergent and customized bricks with different granularities may lead to varying extents of redundancy and collision of capabilities and knowledge. Therefore, detecting the underlying collision and redundancy between bricks is necessary for constructing customized bricks that effectively align with emergent bricks, which makes it possible to achieve optimal performance at minimal cost.

\textbf{Utilization}
The other perspective of avoiding collision and reducing redundancy lies in the joint operations of emergent and customized bricks. As mentioned in $\S\ \ref{router}$, emergent bricks tend to be selected by routing due to their relatively limited number. In contrast, customized bricks are retrieved to augment the current model with various external capabilities. Currently, the routing and retrieval processes of emergent and customized bricks are typically conducted independently, ignoring the potential collaboration. Jointly routing and retrieving emergent and customized bricks can benefit mutually, optimizing collision detection and selection efficiency. In addition, compared with integrating bricks at the model level, stitching emergent and customized bricks with varying granularities may improve the efficiency and reusability of configurable LLMs.

\subsection{Brick Construction Protocols}
\label{protocol}

Configurable LLMs transform the paradigm of LLM alignment and adaptation from a full-parameter training approach to one focusing on the construction and updating of a limited number of bricks. However, most existing algorithms for brick construction and updating, while not requiring the entire LLMs to be updated, still necessitate involving all LLM parameters in the error backpropagation to compute the gradients of bricks. This means that the brick training process demands substantial computational resources. This leads to a contradiction where the bricks offer computational benefits for inference while still being constrained by traditional, resource-intensive training methods.

Efficient brick construction has emerged as a critical challenge. In configurable LLMs, different bricks exchange information through continuous hidden vectors. The primary objective in constructing a brick is to enable it to comprehend the input hidden vectors and generate output hidden vectors that are information-rich and understandable by subsequent modules. It implies that if one can effectively define the input and output hidden vector spaces of a brick, its construction can be independent of the massive parameters of the original LLM. 
To this end, \cite{DBLP:journals/corr/abs-2302-04870} and \cite{DBLP:journals/corr/abs-2212-04129} make preliminary exploration by introducing a small auxiliary model that serves as an emulator for the original LLM. The emulator shares the same brick structure as the original LLM and the hidden vector spaces for inter-brick communication are also pre-aligned with LLM. Each brick in the emulator has significantly fewer parameters compared to its counterpart in the original LLM. Therefore, we can utilize the emulator to construct functional bricks efficiently, which can be directly applied to the original LLM.
Here, the emulator can be regarded as the brick construction protocol designed for LLMs, and bricks built following the protocol can be seamlessly integrated into the LLM. 

A unified and efficient brick construction protocol holds immense potential for the collaborative construction of future LLMs, enabling a paradigm akin to open-source code repository development. The protocol allows multiple developers to engage in collaborative LLM training, and brings two key benefits:
(1) \textit{Protection of Data Privacy}: Developers can utilize their private data to construct high-quality bricks without exposing privacy to a central training process. 
(2) \textit{Distributed Model Training}: Each developer can develop and share bricks based on a unified protocol, without the need for gradient or data communication between different computational nodes.

Despite these advantages, developing more effective and efficient protocols still requires considerable future efforts to fully realize the potential of this collaborative approach:

(1) \textit{Universal Protocols}: The emulator-based approach is usually limited to the inherent structure of the emulator, restricting its applicability to the construction of specific types of bricks. For instance, existing studies develop emulators that preserve only the layer-wise structure of origin LLMs, tailored for bricks that are inserted between transformer layers. 
However, due to the loss of intra-layer vector spaces, the emulator falls short when it comes to constructing bricks within a layer, such as prefixes inserted in attention mechanisms~\citep{DBLP:conf/acl/LiL20}. Therefore, how to design universal protocols suitable for multiple types of bricks remains a great challenge.
(2) \textit{Effective Protocols}: Existing emulators created via pruning or distillation, despite their smaller parameter scale, struggle to accurately represent the vector space of LLMs, thus leading to a performance loss. Therefore, a key focus of future research lies in enhancing the ability of small emulators to better approximate the vector space of LLMs.

\subsection{Evaluation of Configurable Foundation Models}
\label{safety}


Configurable foundation models consist of various functional bricks. It introduces a fresh methodology to evaluate models from the perspective of bricks for existing metrics. Besides, the modular structure and further operations for bricks require evaluating the brick decomposition performance, i.e., whether the bricks can effectively support complex brick operations. 

\textbf{Evaluation from the Perspective of Bricks}
Traditional evaluation methods and metrics usually treat the given LLMs as black-box systems, assessing the ability to generate responses that meet predefined requirements given specific instructions. However, such evaluations typically employ coarse-grained metrics that fail to capture the fine-grained performance. For instance, many efforts use quality scores of model responses or the winning rate against reference responses as metrics for LLM alignment. Such methods can provide coarse-grained performance evaluation but cannot measure the performance in intention understanding, multi-step reasoning, and other fine-grained capabilities.
Configurable foundation models provide a new perspective to model evaluation, allowing us to shift from end-to-end black-box evaluation to brick-by-brick capability evaluations. This approach enables more precise identification of model shortages and directly updating bricks in urgent need of improvement. In this regard, some researchers have begun to explore the brick evaluation. Such as, \citet{DBLP:conf/emnlp/GevaCWG22} examine the functionality of neurons in LLMs, and find that some neurons are responsible for generating undesired toxic language, and deactivating them can achieve effective detoxicity.

\textbf{Evaluation for the Brick Decomposition}
The configurable foundation model also introduces new requirements for model evaluation, particularly regarding whether the bricks within the model can effectively support the diverse brick operations. In this context, we propose the following evaluation metrics for configurable foundation models: 
(1)~Sparsity: A configurable foundation model, during inference, only needs to select a small subset of bricks with relevant functionalities for computation, thereby enhancing inference efficiency. Thus, the goal is to achieve high performance with the least possible number of bricks engaged for given instructions. The fewer bricks required, the more efficient and sparse the model is considered to be.
Some existing efforts focus on actively enhancing model sparsity to minimize the parameters involved for each instruction, thereby improving the computational efficiency of LLMs~\citep{song2024prosparse,szatkowski2024sadmoe}.
(2)~Coupling: As a core concept in software development, decoupling aims to isolate the code that performs a specific task from the code that performs another task~\citep{raghavan2012software}. Indeed, decoupling makes the code more maintainable, reusable, and easier to test~\citep{mo2016decoupling}, which is also important for LLMs. In a configurable foundation model, different bricks are required to be combined to achieve complex capabilities. Besides, the updating and growing operation also needs to update the brick parameters for continual learning. These operations require low-dependency relationships between different bricks, allowing each brick to cooperate with others and be reused across various scenarios multiple times. Additionally, low coupling ensures that changes in one module do not adversely affect the performance of others. In this regard, some efforts have been made to construct task-decoupled knowledge plugins, enabling the reuse of knowledge encoding across various tasks~\citep{DBLP:conf/acl/ZhangZLWYXHLLSZ23,DBLP:conf/acl/XiaoZHCLLLLCS23}.

\subsection{Efficient Brick Computing Frameworks}
\label{compute-framework}

Decomposing LLMs into bricks allows for computation with only a fraction of the parameters, thereby reducing computational load. However, this approach also introduces additional time for brick selection and memory scheduling. Moreover, decoupling the computations of different bricks shows potential for distributed training.
Consequently, to enhance the practicality of configurable foundation models, it is crucial to develop corresponding sparse and heterogeneous computing operators. These research directions are vital for optimizing the efficiency and effectiveness of configurable LLMs, making them more feasible and scalable in diverse computational environments.

\textbf{Sparse Operator}
We have introduced numerous bricks. However, to handle specific inputs, we do not need to use all bricks. If only the bricks that are most effective for specific inputs are used and other bricks are ignored, the computational cost can be significantly reduced. 
However, if the size of the brick is small, sequentially calling multiple bricks will result in a lower utilization rate of CUDA computing units. Therefore, \citet{DBLP:conf/acl/ZhangL00S022} and \citet{DBLP:conf/osdi/CuiHOWZM00XQZ0T23} cluster bricks that are frequently used simultaneously and fuse them into one kernel for parallel execution. \citet{DBLP:conf/icml/LiuWDZY0S0TRC23} implements sparse operators dynamically based on actual input to aggregate bricks.
Given an input, whether a brick is suitable or not generally needs to be judged based on the actual activation value after the calculation of the brick. To avoid these calculations, the above solutions need to perform statistical analysis on a large amount of data beforehand to quickly predict the brick selection plan based on the input.
However, during the training process, the parameters of the brick continue to change, and the applicability for input also changes dynamically. The above solutions that require prior statistical analysis can only be applied to the inference stage after the model is fixed, how to apply them to the training stage remains a challenging problem.

\textbf{Heterogeneous Operator}
Due to the independence between bricks, bricks can be distributed across different machines for collaborative training and inference.
Gshard~\citep{DBLP:conf/iclr/LepikhinLXCFHKS21} and Switch Transformers~\citep{DBLP:journals/jmlr/FedusZS22} leverage the MoE architecture~\cite{DBLP:conf/iclr/ShazeerMMDLHD17} to distribute multiple bricks across different GPUs for parallel pre-training, efficiently scaling up the model size. In particular, the parameter count of the Switch Transformer has reached the trillion level, which is far beyond the size of single-module models. Recent work~\citep{DBLP:journals/corr/abs-2103-13262,DBLP:conf/ppopp/HeZAWLSL22,DBLP:conf/usenix/ZhaiHMZZZ23,DBLP:journals/corr/abs-2206-03382,DBLP:journals/corr/abs-2211-15841} has attempted to solve the problem of load imbalance across different GPUs during MoE training by optimizing brick allocation strategies and scheduling schemes.

During inference, we can place core bricks on servers and custom bricks on user machines~\citep{DBLP:conf/usenix/ZhouWZS22,DBLP:conf/acl/CuiLDHLS23}. In this way, users can conveniently adjust the sub-functions of the model according to their personalized needs, while leaving the core, general, and computation-intensive modules to be computed by the model service provider. On the other hand, when the personalized modules and core modules are placed on different machines, more personalized problems such as how to avoid privacy concerns when transfering private data over the network and how to reduce the inference latency caused by cross-machine communication remain to be solved.

\subsection{Multi-Model Cooperation System}
\label{multi-model}




%

In configurable LLMs, individual bricks collaborate to complete complex instructions. However, building bricks from scratch requires the collection of massive data and consumes significant computational resources. In the rapidly evolving AI community, numerous researchers have open-sourced various pre-trained models with unique capabilities, such as image generation, speech recognition, etc. Reusing and combining these models as model-level bricks can cost-effectively construct a multimodal system capable of handling complex instructions~\citep{DBLP:journals/corr/abs-2401-08525}.

As discussed in $\S\ \ref{plugin}$ and $\S\ \ref{combination}$, there have already been many attempts to combine multiple models to achieve composite capabilities. For example, different modality models are concatenated to achieve multimodal understanding and generation, or different models act as agents that interact with each other through human-readable signals. However, implementing a multi-model cooperation system still faces the following challenges:

\textbf{Scalable Cooperation} Most current works focus on the cooperation of a limited number of models and adapt each model to the entire system, often requiring training of the whole system, which incurs significant overhead. Therefore, designing a highly scalable multi-model framework is an important future direction, which enables the system to efficiently integrate an independently trained model into this multi-model system.

\textbf{Effective Scheduling and Communication} A complex instruction requires different models to perform their duties and coordinate with each other, necessitating that the multi-model system effectively schedules different models and ensures efficient information communication between them. Using human-readable signals for information exchange among different models often leads to information loss. However, the representational spaces of different models vary significantly, and direct interaction using intermediate representations makes it difficult for models to understand each other. To effectively address this issue, one possible approach is to introduce an intermediary model that acts as a bridge and information relay between the different models. Another possible approach is to design a unified intermediate representation form for interactions between different models. Overall, achieving efficient collaboration in complex multi-model systems is an important topic that warrants further research.

\section{Conclusion}
In this paper, we explore configurable foundation models that consist of emergent bricks generated during pre-training and customized bricks created during post-training. We first describe how the bricks constituting a foundational model are trained and further discuss the capabilities of bricks at different granularities. We summarize the advantages of decomposing the foundation models from a modular perspective, including computational efficiency, parameter reusability, traceable results, sustainable capability growth, and optimization for distributed computing. Furthermore, we define four fundamental operations for configurable bricks, including routing and retrieval, brick combination, brick growing, and brick updating. These four operations enable the completion of complex instructions even when each brick is responsible for a single capability. Finally, we discuss the open problems and challenges that remain unresolved for configurable foundation models. We hope this paper will stimulate further research to construct more efficient and scalable foundation models from a modular perspective.

\section*{Contributions}
The contributions of the authors are listed as follows: Chaojun Xiao, Zhengyan Zhang, Xu Han, Zhiyuan Liu, and Maosong Sun initiated and organized the research. Chaojun Xiao drafted the abstract. Chaojun Xiao, Zhengyan Zhang, and Xu Han drafted the introduction. Zhengyan Zhang, Feng Yao, and Xiaozhi Wang drafted $\S\ \ref{emergent}$. Chaojun Xiao and Xiaozhi Wang drafted $\S\ \ref{plugin}$. Chenyang Song and Shuo Wang drafted $\S\ \ref{granularity}$. Chaojun Xiao and Yuge Tu drafted $\S\ \ref{benefit}$. Yufei Huang drafted $\S\ \ref{router}$. Chaojun Xiao drafted $\S\ \ref{combination}$. Yingfa Chen drafted $\S\ \ref{updating}$. Chenyang Song drafted $\S\ \ref{growing}$. Chaojun Xiao, Dazhi Jiang, and Chenyang Song conducted experiments for $\S\ \ref{experiments}$. Feng Yao drafted $\S\ \ref{correlation}$. Chaojun Xiao drafted $\S\ \ref{protocol}$. Guanyu Lin and Chaojun Xiao drafted $\S\ \ref{safety}$. Weilin Zhao drafted $\S\ \ref{compute-framework}$. Chaojun Xiao drafted $\S\ \ref{multi-model}$. Chaojun Xiao drafted the conclusion. Jingbo Shang, Huimin Chen, Yankai Lin, Zexuan Zhong, Ao Zhang, and Chenglei Si proofread the paper and provided valuable feedback on the paper structure. Khai Hao Moo and Chenyang Zhao proofread the paper and provided valuable suggestions for grammar correction.

\bibliographystyle{citation}
\bibliography{custom}

\end{document}